\documentclass[a4paper,fleqn]{cas-sc}
\usepackage[numbers]{natbib}
\usepackage{makecell}
\usepackage{pdflscape}
\usepackage{longtable}
\usepackage{csvsimple-l3}
\usepackage{subcaption}
\usepackage{pdfpages}
\usepackage{comment}
\usepackage{threeparttable}

\begin{document}

\title[mode = title]{Unsupervised Deep Generative Models for Anomaly Detection in Neuroimaging: A Systematic Scoping Review}
\shorttitle{Unsupervised Deep Generative Models for Anomaly Detection in Neuroimaging} 
\shortauthors{Y. Mahé et~al.}

\author[1,2]{Youwan Mahé}[orcid=0009-0006-9289-7178]
\cormark[1]
\ead{youwan.mahe@inria.fr}

\author[1,3]{Elise Bannier} 
\author[1,4,5]{Stéphanie Leplaideur}

\author[6]{Elisa Fromont}
\fnmark[1]

\author[1]{Francesca Galassi}[orcid=0000-0002-8788-2856]
\cormark[2]
\fnmark[1]
\ead{francesca.galassi@inria.fr}

\cortext[cor1]{Corresponding author} 
\cortext[cor2]{Principal corresponding author}

\fntext[fn1]{EF and FG contributed equally as co–last authors.}

\affiliation[1]{organization={Univ Rennes, Inria, CNRS, Inserm, IRISA UMR 6074, Empenn}, city={Rennes}, country={France}}

\affiliation[2]{organization={Siemens Healthineers}, city={Courbevoie}, country={France}}
\affiliation[3]{organization={CHU Rennes, Radiology Department}, city={Rennes}, country={France}}
\affiliation[4]{organization={CHU Rennes, Physical Medicine and Rehabilitation Department}, city={Rennes}, country={France}}
\affiliation[5]{organization={Centre de Kerpape}, city={Ploemeur}, country={France}}
\affiliation[6]{organization={Univ Rennes, Inria, CNRS, IRISA}, city={Rennes}, country={France}}

\begin{abstract}
Unsupervised anomaly detection (UAD) based on deep generative modelling has been increasingly investigated for identifying pathological brain abnormalities without requiring voxel-level annotations. By learning the distribution of healthy anatomy and generating pseudo-healthy reconstructions, these methods aim to localise deviations without requiring pathology-specific examples during training. Despite rapid methodological development - from autoencoders and variational autoencoders to generative adversarial networks and diffusion-based models - a structured synthesis of their application in structural neuroimaging is lacking.

We conducted a PRISMA-ScR-guided scoping review of studies published or made available online between January 2018 and the search date (17 December 2025) that applied unsupervised deep generative models to anomaly detection in brain MRI (and, less frequently, CT). Thirty-three studies met the inclusion criteria. Methods were categorised by architectural family, and reported performance was synthesised across major pathology groups, with segmentation (Dice) and detection metrics (AUROC, AUPRC) disaggregated by evaluation level (voxel, slice, subject). For transparency, we also summarised dataset characteristics, dimensionality (2D vs.\ 3D), and thresholding strategies.

Reported performance varied substantially across studies and pathologies. Dice scores generally ranged from approximately 0.30 to 0.75, while voxel-level AUROC values spanned roughly 0.58 to 0.97. Performance was typically higher for large focal lesions, such as brain tumours, than for small or diffuse abnormalities including white-matter hyperintensities. Although performance ranges overlapped substantially across architectural families, differences in evaluation level, thresholding strategy, dataset composition, and other experimental conditions preclude reliable attribution of cross-study performance differences to pathology-related factors or architectural design.

Overall, unsupervised generative approaches show potential for anomaly localisation without requiring pathology-specific examples during training. However, methodological heterogeneity, limited external validation, and sensitivity to dataset characteristics remain major barriers to clinical translation. Emerging paradigms - including anatomy-aware modelling, diffusion-based frameworks, and alternative normative evaluation metrics - seek to improve robustness and support clinical translation.
\end{abstract}


\maketitle

\section{Introduction}

Brain magnetic resonance imaging (MRI) plays a central role in the diagnosis, monitoring, and prognosis of neurological disease. Structural MRI enables high-resolution \textit{in vivo} visualisation of abnormalities such as tumours, vascular infarcts, white-matter hyperintensities, demyelinating lesions, traumatic injury, and regional atrophy~\citep{Vemuri2022, VillanuevaMeyer2017}. Their appearance varies substantially across imaging sequences and pathology types~\citep{Crimi2021}. Accurate segmentation is essential for deriving quantitative biomarkers - including lesion load, spatial distribution, and longitudinal volumetric change - that support clinical decision-making and research. However, manual lesion detection and segmentation by expert radiologists are time-consuming, require specialised expertise, and are subject to inter- and intra-rater variability \citep{Darrault2025, walsh2023expert}. The development of robust automated tools is therefore a major priority.

Before the rise of deep learning, automated lesion detection relied on classical image-processing and statistical modelling approaches, including Gaussian mixture models, fuzzy c-means clustering, and Markov random fields \citep{Gonzalez2007, GarciaLorenzo2013}. While effective in controlled settings, these techniques often relied on hand-crafted features or explicit modelling assumptions and were sensitive to anatomical variability and acquisition differences \citep{Xu2024}. The advent of deep learning transformed medical image analysis by enabling multi-scale feature representations to be learnt directly from data \citep{Litjens2017}. In neuroimaging, convolutional architectures such as U-Net \citep{Ronneberger2015} and its derivatives, including nnU-Net \citep{Isensee2024}, have achieved state-of-the-art performance on curated benchmarks. Large public datasets such as BraTS for brain tumours \citep{Menze2015} and MSSEG for multiple sclerosis \citep{Commowick2021} have facilitated this progress.

Despite these advances, supervised models exhibit important limitations, particularly when deployed in clinical practice. Their development relies heavily on large voxel-level annotated datasets and typically follows a closed-set assumption, whereby the pathological patterns encountered at inference are expected to be represented in the training data. Consequently, performance may deteriorate when models encounter previously unseen pathological presentations. In addition, distribution shifts arising from differences in acquisition protocols, scanners, or patient populations may further compromise generalisation. These limitations motivate approaches that do not depend on pathology-specific training examples.

Unsupervised anomaly detection (UAD) offers a complementary paradigm. Rather than learning from labelled pathological examples, UAD methods aim to model the distribution of healthy anatomy and detect deviations from this normative representation. In reconstruction-based generative UAD, models are typically trained exclusively on healthy brain scans; at inference, a potentially pathological image is reconstructed as a pseudo-healthy image, and spatial discrepancies between the input and its reconstruction indicate candidate abnormalities. This healthy-only training paradigm does not require pathological examples or voxel-level lesion annotations for model training, providing a framework for identifying candidate abnormalities that were not represented in the training data. The importance of robust detection under open-world conditions has recently been highlighted by benchmarks such as NOVA \citep{Bercea2025-2}, which evaluates models on a diverse collection of rare brain MRI pathologies and demonstrates the difficulty of anomaly detection in such settings.

Over the past decade, deep generative modelling for anomaly detection has evolved rapidly. Early approaches employed autoencoders and variational autoencoders (VAEs) to learn representations of healthy anatomy and identify abnormalities through reconstruction errors \citep{Zimmerer2019}. Generative adversarial networks (GANs) subsequently incorporated adversarial learning to model the distribution of healthy images and generate realistic reconstructions \citep{Schlegl2019}. More recently, denoising diffusion probabilistic models (DDPMs) \citep{Ho2020} and related diffusion-based frameworks have introduced iterative denoising as an alternative generative mechanism for pseudo-healthy reconstructions \citep{Pinaya2022-3, Bercea2023}. These methodological advances have expanded the landscape of unsupervised neuroimaging anomaly detection.

Several reviews have addressed subsets of this literature, including autoencoder- and diffusion-based anomaly detection \citep{Baur2021-1, BerceaReview2025}, GAN applications in medical imaging \citep{Wang2023}, and broader surveys of generative models in healthcare \citep{Pang2021, Tschuchnig2022}. However, to our knowledge, no prior review has systematically synthesised unsupervised deep generative models for structural neuroimaging from a pathology-stratified perspective while explicitly examining segmentation and detection metrics, evaluation levels (voxel, slice, and subject), thresholding strategies, and dataset characteristics. Reported performance varies substantially across pathology types, evaluation protocols, and dataset characteristics, complicating cross-study comparison and interpretation.

In this PRISMA-ScR-guided scoping review, we systematically analyse unsupervised deep generative models for anomaly detection in structural neuroimaging published or made available online between 2018 and the search date (17 December 2025). We categorise methods into four principal architectural families - autoencoders, variational autoencoders, generative adversarial networks, and diffusion-based models - and synthesise reported performance across major pathology groups, including brain tumours, stroke, multiple sclerosis, and white-matter hyperintensities. Beyond characterising these architectural families, we examine how reported metrics vary according to dimensionality (2D vs.\ 3D), dataset composition, thresholding strategies, and evaluation level. Finally, we discuss methodological limitations of reconstruction-based approaches and highlight emerging paradigms aimed at improving robustness, interpretability, and clinical relevance.

\subsection{Review questions}
This scoping review addresses the following questions:
\begin{enumerate}[(\roman*)]
    \item Which unsupervised deep generative model families (AE, VAE, GAN, and diffusion-based models) have been applied to anomaly detection and/or segmentation in structural neuroimaging since 2018, and which specific pathologies do they target?
    \item How are anomaly detection and segmentation evaluated in terms of evaluation level, metrics, and thresholding strategy, and how do reported results vary across major pathology groups?
    \item What methodological design variations (e.g., 2D vs.\ 3D processing, patching or masking strategies, and pretraining) are described in the literature, and how does reported performance vary across these methodological designs?
    \item What emerging paradigms aim to address current limitations of generative UAD?
\end{enumerate}


\section{Methods}
\subsection{Background: generative modelling for unsupervised anomaly detection} 
\label{Background}

Generative models aim to learn the distribution of healthy brain images, thereby capturing normative anatomical variability. When an explicit probabilistic formulation is available, it can be written as approximating the data distribution $p_{\text{data}}(\mathbf{x})$ with a distribution $p_\theta(\mathbf{x})$ induced by a parameterised mapping. In neuroimaging UAD, models are trained on healthy brain images only, so the learned distribution represents normative anatomy rather than pathology.

Many generative approaches introduce a latent variable $\mathbf{z} \in \mathbb{R}^d$, drawn from a simple prior distribution $p(\mathbf{z})$, typically a standard multivariate Gaussian $\mathcal{N}(\mathbf{0}, \mathbf{I})$, and a neural network $G_\theta$ that maps latent codes to image space:

\begin{equation}
\mathbf{z} \sim p(\mathbf{z}) = \mathcal{N}(\mathbf{0},\mathbf{I}), \qquad 
\mathbf{x} = G_\theta(\mathbf{z}), \qquad 
\mathbf{x} \sim p_\theta(\mathbf{x}).
\end{equation}

When an explicit encoder $E_\phi$ is available (e.g., autoencoders and VAEs), a test image $\mathbf{x}_{\text{test}}$ is mapped to latent space as 
\begin{equation}
\mathbf{z} = E_\phi(\mathbf{x}_{\text{test}}), \qquad
\hat{\mathbf{x}} = G_\theta(\mathbf{z}).
\end{equation}
For models without an encoder (e.g., classical GANs or some diffusion-based
methods), a latent representation must instead be obtained through an inversion
or optimisation procedure:
\begin{equation}
\mathbf{z}^\ast = \arg\min_{\mathbf{z}} \;
\mathrm{dist}\big(\mathbf{x}_{\text{test}}, G_\theta(\mathbf{z})\big), \qquad
\hat{\mathbf{x}} = G_\theta(\mathbf{z}^\ast),
\end{equation}
where $\mathrm{dist}(\cdot,\cdot)$ denotes a similarity or dissimilarity measure
(e.g., $\ell_1$, $\ell_2$, SSIM).

Reconstruction-based UAD relies on the premise that pathological features are reconstructed less faithfully than healthy anatomy, such that discrepancies between the input image and its pseudo-healthy reconstruction can indicate candidate abnormalities. A spatial anomaly map can therefore be defined as
\begin{equation}
\mathbf{r} = \mathrm{dist}\big(\mathbf{x}_{\text{test}}, \hat{\mathbf{x}}\big),
\end{equation}
where $\mathrm{dist}(\cdot,\cdot)$ denotes a spatially resolved discrepancy measure applied between the input and reconstructed images. The resulting anomaly map highlights regions in which the observed anatomy deviates from its reconstructed normative representation, without requiring voxel-level pathological annotations during model training.

Despite substantial architectural differences between autoencoders, variational autoencoders, generative adversarial networks, and diffusion models, reconstruction-based anomaly detection underpins the majority of generative UAD approaches reviewed in this work. Subsequent sections examine how architectures, dimensionalities, and training strategies vary across studies and how these methodological differences are reported in relation to reconstruction quality and anomaly detection performance.

\subsection{Information sources and search strategy}
This review was conducted in accordance with the PRISMA-ScR (2018) guidelines for scoping reviews \citep{Tricco2018}.  We searched six information sources: PubMed, Web of Science, ScienceDirect, Springer Nature Link, IEEE Xplore, and ArXiv up to 17 December 2025, including early-access articles and preprints. Boolean queries combined terms related to \textit{unsupervised anomaly detection}, \textit{neuroimaging} (MRI or CT), and \textit{deep learning}, as summarised in Table~\ref{tab:prompts}. Reference lists of relevant articles and reviews were also screened. Searches were limited to articles published in English.

\begin{table*}[pos=ht]
    \caption{Boolean queries used for literature searching.}
    \label{tab:prompts}
    \begin{tabular*}{0.9\textwidth}{@{} l L l @{}}
    \toprule
    \textbf{Database} & \textbf{Query} & \textbf{Date} \\
    \midrule
    PubMed & \makecell[l]{(Anomaly AND Unsupervised AND Brain)} \\AND (MRI OR CT) \\ AND (Machine Learning OR Deep Learning) & Dec 17 2025 \\
    \midrule
    Web Of Science & \makecell[l]{((TS=Anomaly) AND (TS=unsupervised) \\ AND (TS=brain)) AND ((TS=MRI) \\ OR (TS=CT)) AND ((TS=Machine Learning) \\ OR (TS=Deep Learning))} & Dec 17 2025 \\
    \midrule
    Science Direct & \makecell[l]{\textit{Unsupervised Anomaly Brain MRI CT Machine Learning} \\ Filter : Research Article} & Dec 17 2025 \\
    \midrule
    ArXiv & \makecell[l]{Unsupervised AND Anomaly AND Brain \\ AND "Deep learning", in Computer Science (cs)} & Dec 17 2025 \\
    \midrule
    IEEE Xplore & \makecell[l]{Unsupervised AND Anomaly AND Brain AND "Deep Learning"} & Dec 17 2025 \\
    \midrule
    Springer Nature Link & \makecell[l]{(Anomaly AND Unsupervised AND Brain) \\ AND (MRI OR CT) \\ AND (Machine Learning OR Deep Learning) \\ AND ("Conference Paper" OR "Research Article") \\ AND ("Computer Vision" OR "Machine Learning")} & Dec 17 2025 \\
    \bottomrule
    \end{tabular*}
\end{table*}

\subsection{Eligibility and screening}

Studies were eligible if they: 
\begin{itemize}
    \item proposed or evaluated an unsupervised deep learning method for anomaly detection and/or lesion segmentation in structural neuroimaging (MRI or CT), trained to model normative anatomy without using real pathological labels in the training objective;
    \item reported at least one quantitative evaluation metric relevant to detection (e.g., AUROC, AUPRC) or segmentation (e.g., Dice);
    \item used real human imaging data from public datasets or institutional cohorts;
    \item were published or made available online between January 2018 and the predefined search deadline (17 December 2025), irrespective of the formal publication year.
\end{itemize}

Unsupervised learning can be characterised in multiple ways. A narrow definition restricts it to methods that lack any explicit label-driven training objective, thereby excluding self-supervised approaches~\citep{Dosovitskiy2014}. In this review, we adopt a pragmatic definition aligned with anomaly detection in medical imaging: we include methods trained without using real pathological labels (neither voxel-level segmentations nor subject-/case-level diagnostic labels) in the training objective. This encompasses approaches that use proxy or self-supervised objectives to model the distribution of healthy data (e.g., masked autoencoding, lesion synthesis, or discriminative proxy tasks)~\citep{Goodfellow2016}. In Table~\ref{tab:method_summary}, methods employing such proxy or self-supervised objectives are denoted with a $^\mathcal{P}$.

Studies were excluded if they employed rule-based or non-deep learning approaches; fully supervised or semi-supervised methods requiring pathological annotations; non-generative methods (except where discussed contextually); non-neuroimaging applications; animal-only or synthetic-only data; review papers; and non-research formats (e.g., abstracts, editorials).
For transparency, self-supervised non-generative and hybrid approaches are discussed in a separate contextual subsection but are not included in the main quantitative synthesis because they do not satisfy the generative modelling inclusion criteria. Although some applications use deep generative models for functional or diffusion imaging (e.g., FDG-PET, diffusion tensor imaging, or functional MRI)~\citep{Hassanaly2024-ex, Solal2023-ex,Munoz-Ramirez2022-ex,Sathyanarayanan2025-ex}, we restricted the primary synthesis to structural imaging to ensure comparability across studies. Similarly, we excluded studies that did not report voxel-, slice-, or subject-level quantitative performance metrics.

Search results were imported into Rayyan \citep{Ouzzani2016} for duplicate removal and blinded screening by two reviewers. Titles and abstracts were screened first, followed by full-text assessment of potentially relevant studies. Disagreements were resolved through discussion until consensus was reached.

Bibliographic records were retrieved using the export functions of each database (CSV or BibTeX) and imported into Rayyan for record matching and management. For arXiv, where no export function is available through the web interface, we used the API via custom Python scripts to obtain publication metadata. Full-text articles were accessed through institutional subscriptions where available.

Data extraction was performed independently by two reviewers using a structured extraction form developed for this review. Extracted variables included: publication year, model architecture, imaging modality, dimensionality (2D or 3D), training and test datasets, pathology type, evaluation level (voxel, slice, or subject), thresholding strategy, and reported quantitative performance metrics (Dice, AUROC, AUPRC). Extracted data were tabulated and used for qualitative and descriptive quantitative synthesis.

\subsection{Quality appraisal and integrity screening}
Although formal risk-of-bias assessment is not routinely undertaken in scoping reviews, we performed targeted appraisals to contextualise potential limitations of the included evidence. First, two reviewers (YM, FG) independently appraised the datasets used across the included studies, focusing on characteristics relevant to potential sources of bias and generalisability, including sample size, population characteristics, acquisition setting, imaging modality, and centre composition. These dataset-level characteristics are summarised in Table~\ref{tab:dataset_summary}. Second, we performed a study-level methodological appraisal focusing on single-centre evaluation, internal evaluation, post-hoc threshold optimisation, and public code availability, as summarised in Table~\ref{tab:bias}. Additionally, in cases of ambiguity, studies were screened for potential research integrity concerns using the \textit{Problematic Paper Screener}~\citep{Cabanac2022}.


\section{Results}
\subsection{Study selection}
\begin{figure}
    \centering
    \includegraphics[width=\linewidth]{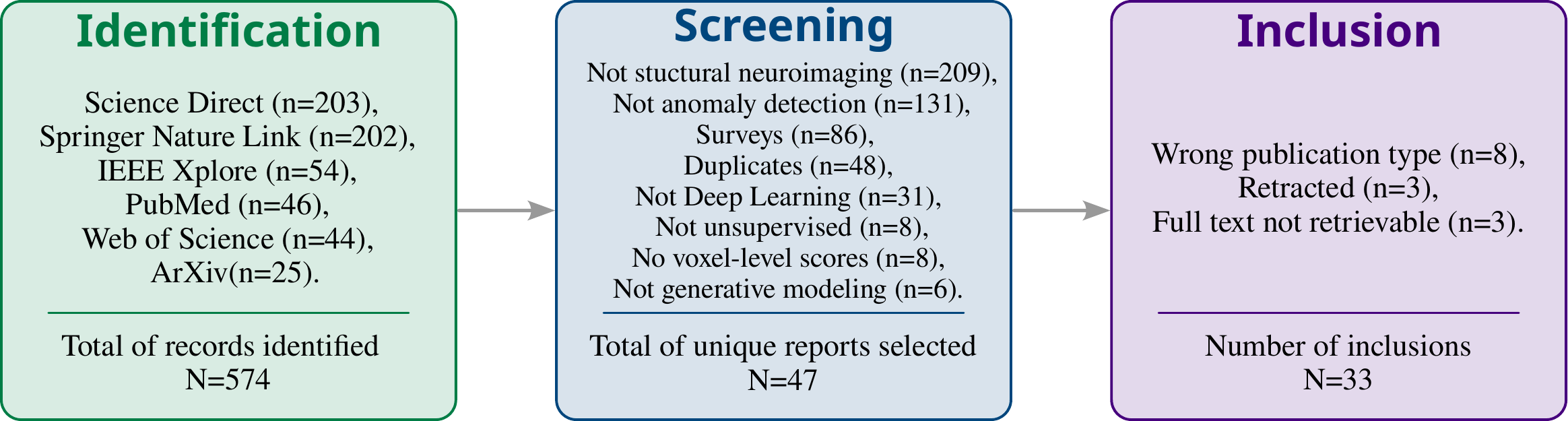}
    \caption{PRISMA flow diagram for scoping reviews, including database and register searches.}
    \label{fig:prisma-flowchart}
\end{figure}

The initial search identified 574 records. After duplicate removal and title/abstract screening, 527 records were excluded according to predefined eligibility criteria (e.g., out of scope, non-neuroimaging, non-deep learning, rule-based methods, review articles). Forty-seven full-text articles were assessed for eligibility, of which 14 were excluded (e.g., absence of structural imaging, lack of quantitative metrics, non-research format). A total of 33 studies were included in the final synthesis (Fig.~\ref{fig:prisma-flowchart}). 
The included studies were published or made available online between January 2018 and 17 December 2025, including early-access articles and preprints, providing an eight-year overview of unsupervised deep generative models applied to structural neuroimaging.

\subsection{Study characteristics}
Across the 33 included studies, four main architectural families were identified: autoencoders (AEs), variational autoencoders (VAEs), generative adversarial networks (GANs), and diffusion-based models. Although standard AEs do not explicitly define a probabilistic generative model, they are widely used as reconstruction-based normative baselines and are therefore included in this synthesis. Proxy or self-supervised pretext objectives were included when used to support pseudo-healthy reconstruction (e.g., masking/inpainting or structured corruption). In total, the 33 studies comprised 9 autoencoder-based approaches, 9 variational autoencoders, 4 GAN-based methods, and 11 diffusion-based models.

MRI was the primary imaging modality (T1-w, T1-w Gd, T2-w, and fluid-attenuated inversion recovery (FLAIR)), with occasional use of computed tomography. Most methods processed 2D slices, although a subset employed volumetric 3D architectures, particularly among recent AE/VAE studies and selected GAN and diffusion approaches. Although some generative models have been applied to non-structural modalities such as FDG-PET for Alzheimer's disease detection~\citep{Hassanaly2024-ex, Solal2023-ex}, this review focused on structural imaging to ensure methodological comparability across studies.

\subsection{Evaluation metrics}
Most studies reported either segmentation or detection performance. Segmentation accuracy was assessed using the Dice similarity coefficient (Dice score), defined as
\begin{equation}
\text{Dice} = \frac{2|X \cap Y|}{|X| + |Y|},
\label{eq:dice}
\end{equation}
where $X$ is the predicted segmentation and $Y$ the reference annotation. Dice scores depend not only on the quality of the anomaly map but also on the thresholding strategy used to binarise residual maps. Threshold selection varied substantially across studies. In several cases, thresholds were retrospectively chosen to maximise Dice on the test set, yielding optimistic upper-bound estimates that may not reflect clinically deployable performance. Dice is also sensitive to lesion size and can disproportionately penalise small or sparse abnormalities.
Detection was formulated as a binary classification problem (pathological vs.\ non-pathological) and quantified using AUROC and AUPRC. In highly imbalanced voxel-level tasks, AUPRC is often more informative than AUROC because it better reflects precision under class imbalance.
Detection metrics were computed at different evaluation levels (voxel/pixel, slice, subject). Because these correspond to distinct classification tasks, values are not directly comparable across evaluation regimes.

\textit{Comparability across studies.}
Substantial heterogeneity exists in dataset composition, preprocessing pipelines, anomaly-map post-processing, thresholding strategies, and evaluation levels. Accordingly, reported performance values should be interpreted as descriptive indicators within each experimental setting rather than direct head-to-head comparisons across architectures.

\subsection{Datasets and pathologies}
\begin{figure}
    \centering
    \includegraphics[width=\linewidth]{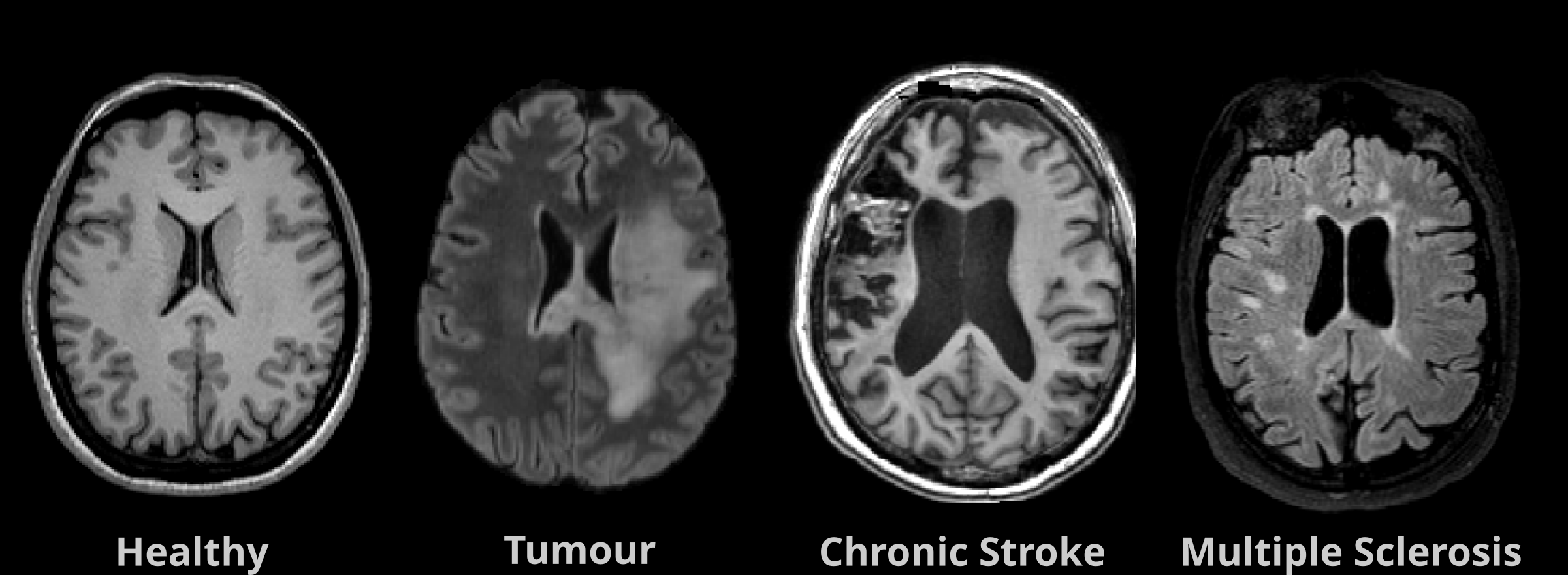}
    \caption{Central axial slices from a healthy brain (IXI, 3D T1), a brain tumour case (BraTS, FLAIR after brain extraction), a chronic stroke case (ATLAS v2.0, 3D T1), and a multiple sclerosis case (MSSEG, FLAIR with fat saturation).}
    \label{fig:AnoExample}
\end{figure}

Performance metrics were reported across a range of publicly available and institutional neuroimaging datasets. Three pathology groups predominated in the literature: brain tumours, multiple sclerosis (MS)/white-matter hyperintensities (WMH), and stroke. Fewer studies addressed neurodegenerative conditions (e.g., Alzheimer's disease, Parkinson's disease), traumatic brain injury, neonatal encephalopathy, or healthy ageing.
Representative examples of central axial slices from the principal pathological datasets are shown in Figure~\ref{fig:AnoExample}. A structured overview of dataset characteristics - including cohort size, imaging modality, and documented biases - is provided in Table~\ref{tab:dataset_summary}. Because dataset composition, acquisition protocols, and pathology prevalence vary substantially across studies, these factors should be considered when interpreting reported performance metrics.

\paragraph{Brain tumours.}  
The majority of tumour studies relied on the \textit{Brain Tumor Segmentation} (BraTS) challenge datasets \citep{Menze2015}, which provide multi-sequence MRI including T1-weighted (T1-w), contrast-enhanced T1-w (T1-w Gd), T2-weighted (T2-w), and FLAIR images. As detailed in Table~\ref{tab:dataset_summary}, BraTS has expanded substantially across releases: BraTS 2012 included 50 cases, BraTS 2015 increased to 253 cases (with a strong high-grade glioma predominance), BraTS 2017-2020 ranged from 477 to 660 cases with expert-revised annotations across multiple centres, and the most recent BraTS 2021-2023 (GLI) editions comprise approximately 2{,}040 multi-institutional cases \citep{Baid2021}. 
A smaller number of studies used alternative cohorts, including the neuroimaging dataset of brain tumour patients from \citet{Pernet2016} (22 cases) and the Brain Tumor Dataset (BTD) from \citet{Cheng2017} (233 cases), which are comparatively smaller and more homogeneous than BraTS. Gliomas, the primary pathology represented in these datasets, typically produce large focal lesions with substantial mass effect. Lesion morphology varies according to tumour grade, and progressive growth within the cranial cavity frequently distorts surrounding anatomical structures. Sequence selection is clinically motivated: T2-w and FLAIR emphasise water content and oedema, whereas T1-w Gd highlights blood-brain barrier disruption through contrast enhancement \citep{Menze2015}.

\paragraph{Stroke.}  
Stroke lesions were primarily represented by the \textit{Ischaemic Stroke Lesion Segmentation} (ISLES) 2015 dataset and the \textit{Anatomical Tracings of Lesions After Stroke} (ATLAS) datasets (versions 1.1/1.2 and 2.0) \citep{HernandezPetzsche2022, Liew2021}, as summarised in Table~\ref{tab:dataset_summary}. ISLES 2015 comprises two subsets: the Sub-acute Ischemic Stroke Lesion Segmentation (SISS) cohort (64 subjects imaged with T1-w, T2-w, FLAIR, and diffusion-weighted imaging (DWI)) and the Stroke Perfusion Estimation (SPES) cohort (50 subjects imaged with T1-w, T2-w, DWI, and perfusion modalities). These cohorts target acute and sub-acute ischaemic stroke, where DWI is most sensitive to early infarction. As stroke progresses, diffusion signal characteristics evolve and lesion conspicuity may decrease, making delineation on structural sequences more challenging \citep{HernandezPetzsche2022}.  
ATLAS v1.1/1.2 includes 229-304 chronic stroke cases acquired as T1-weighted MRI, while ATLAS v2.0 comprises 955 manually segmented T1-w MRI cases collected from 33 cohorts worldwide \citep{Liew2021}. A key limitation of ATLAS is its reliance on T1-w imaging alone, where initial lesion extent blends with subsequent degeneration and tissue loss. Overall, lesion stage (acute vs.\ chronic), available contrasts (multi-sequence vs.\ T1-w only), and lesion appearance vary substantially across datasets.

\paragraph{Multiple sclerosis and white-matter hyperintensities.}
Although multiple sclerosis and white-matter hyperintensities (WMH) arise from distinct pathological processes, they are described jointly here because both commonly manifest as FLAIR-hyperintense white-matter abnormalities and often present relatively low lesion-to-brain volume ratios compared with large focal pathologies such as tumours.
As summarised in Table~\ref{tab:dataset_summary}, MS was primarily evaluated using the \textit{ISBI 2015 Longitudinal MS Lesion Segmentation} dataset (ISBI 2015)~\citep{Carass2017}, the \textit{Multiple Sclerosis Segmentation} dataset (MSSEG)~\citep{Commowick2018, Commowick2021}, and the \textit{MSLUB} dataset~\citep{Lesjak2017}. MSSEG (MICCAI 2016) comprises 53 subjects with four MRI contrasts (T1-w, T1-w Gd, proton density, and FLAIR), while ISBI 2015 includes 82 cases and is subject to mono-site bias. MSLUB is a smaller mono-site cohort of 30 subjects with T2-w, FLAIR, and proton density imaging.
WMH were primarily evaluated using the \textit{WMH Segmentation Challenge} dataset (WMH 2017; 170 subjects; T1-w and FLAIR), which is enriched for older participants and spans multiple scanners, introducing age- and scanner-related heterogeneity \citep{Kuijf2022}.  
Unlike tumours or many stroke lesions, MS plaques and many WMH are relatively small and spatially distributed. They are most conspicuous on FLAIR and T2-w MRI, where they appear as focal white-matter hyperintensities. Their small size and multiplicity present particular challenges for segmentation and anomaly detection.

\paragraph{Traumatic brain injury.}
Traumatic brain injury (TBI) was represented primarily by the \textit{CENTER-TBI} cohort (Table~\ref{tab:dataset_summary}), a large multicentre European dataset comprising $\sim$4{,}500 subjects, with imaging dominated by computed tomography (CT) \citep{Steyerberg2019}. On CT, acute TBI may present as focal hyperdense haemorrhagic lesions (e.g., epidural or subdural haematomas) as well as smaller haemorrhagic foci; however, diffuse axonal injury is often subtle or not directly visible on standard CT, which can limit sensitivity for detecting microscopic injury patterns \citep{Currie2015}.

\paragraph{Neurodegeneration and other conditions.}
A smaller subset of included studies leveraged neurodegenerative cohorts, most often through the \textit{Alzheimer's Disease Neuroimaging Initiative} (ADNI) \citep{Beckett2015} and the \textit{Open Access Series of Imaging Studies} (OASIS-3) \citep{LaMontagne2019}, as summarised in Table~\ref{tab:dataset_summary}. 
ADNI is a large multi-phase public repository including structural MRI and PET imaging from cognitively normal individuals, subjects with mild cognitive impairment, and patients with Alzheimer's disease ($>2{,}000$ participants across phases). OASIS-3 is a longitudinal, single-site cohort comprising 1,098 participants (605 cognitively normal and 493 at-risk or demented individuals), with multi-contrast MRI including T1-w, T2-w, FLAIR, DWI, susceptibility weighted imaging, arterial spin labeling, and resting-state functional MRI. 

\paragraph{Healthy control datasets.}
Across studies, normative training data were sourced either from dedicated healthy cohorts (e.g., IXI, UK Biobank, HCP) or from healthy subsets extracted from mixed pathology datasets (denoted H in Table~\ref{tab:method_summary}). A smaller number of studies relied on in-house institutional cohorts. Explicit reporting of screening procedures and contamination control was inconsistent. Frequently used resources included the IXI dataset\footnote{\url{http://brain-development.org/ixi-dataset/}} (approximately 600 healthy volunteers scanned across three London hospitals; T1-w, T2-w, and proton density MRI), which was widely employed for training autoencoder, VAE, and diffusion models.
Large population-based cohorts were also used, including the UK Biobank (UKB), which provides multi-modal MRI for over 50{,}000 participants, and the Human Connectome Project (HCP; Young Adult cohort, N$\sim$1{,}200; ages 22–35 years)~\citep{VanEssen2012}. While UKB offers unprecedented scale, it remains subject to healthy volunteer and geographical biases \citep{Littlejohns2020}. HCP, although high-quality, reflects a restricted young adult demographic. 
Smaller normative datasets included the Cambridge Centre for Ageing and Neuroscience (Cam-CAN; N$\sim$650)~\citep{Shafto2014}, the Neurofeedback Skull-stripped (NFBS) dataset (125 T1-w scans) \citep{Puccio2016}, and institutional in-house cohorts. A single diffusion-based study used a healthy subset of the FastMRI+ dataset (1{,}001 axial brain MRI scans) \citep{Zhao2022}. The \textit{Medical Out-of-Distribution} (MOOD) challenge~\citep{Zimmerer2020} dataset was used once. Notably, MOOD relies on synthetic anomalies injected into healthy HCP scans, which may introduce synthetic bias and inflate detection performance relative to real-world pathology. Although ADNI and OASIS-3 contain both healthy and pathological subjects, in the reviewed literature, they were primarily used as sources of cognitively normal control scans for training normative models rather than as primary anomaly detection benchmarks. It is important to note that ADNI controls exhibit a documented healthy volunteer bias \citep{Gianattasio2021}, and OASIS-3 reflects single-site demographic characteristics, which may influence normative modelling. 
A single study reported combining smaller healthy subsets of the \textit{individual brain charting (IBC)}, a high-resolution fMRI dataset for cognitive mapping~\citep{IBC2018}, and from a multi-parametric neuroimaging reproducibility study referred here to as \textit{Kirby}~\citep{Landman2011}

\subsection{Synthesis by pathology}
To provide a structured overview, we synthesised findings primarily by pathology while considering architectural family within each pathology group. For segmentation tasks, we report Dice scores, whereas for detection tasks, we report AUROC and AUPRC values, specifying the evaluation level (voxel/pixel, slice, or subject) to avoid misleading cross-study comparisons.
Study characteristics and experimental settings are summarised in five complementary tables. Table~\ref{tab:dataset_summary} describes the datasets (release, size, modality, population, and known biases). Table~\ref{tab:method_summary} summarises the methods (architectural family, training and test datasets, imaging modality, and input dimensionality/processing strategy). Table~\ref{tab:bias} reports the study-level methodological appraisal. Quantitative results are reported in Table~\ref{tab:detection_metrics} for detection performance and Table~\ref{tab:dice_summary} for segmentation performance. 

\begin{landscape}
        \begin{longtable}{lp{4cm}p{2.4cm}p{2.4cm}p{2cm}p{6.5cm}}
\caption{Comprehensive overview of neuroimaging datasets and their characteristics.}
\label{tab:dataset_summary}
\\
\toprule
\textbf{Dataset (Release)} & \textbf{Modality} & \textbf{Pathology} & \textbf{Population} & \textbf{Cases (N)} & \textbf{Known Biases \& Characteristics} \\
\midrule
\endfirsthead
\toprule
\textbf{Dataset (Release)} & \textbf{Modality} & \textbf{Pathology} & \textbf{Population} & \textbf{Cases (N)} & \textbf{Known Biases \& Characteristics} \\
\midrule
\endhead
\midrule
\multicolumn{6}{r}{\textit{Continued on next page}} \\
\midrule
\endfoot
\bottomrule
\endlastfoot

\makebox[0pt][l]{\textbf{Tumour}} & & & & & \\
BraTS '12 & MRI (4 seqs) & Glioma & Adult & 50 & Small sample size.\\
BraTS '15 & MRI (4 seqs) & Glioma & Adult & 253 & Skewed to HGG.\\
BraTS '17-20 & MRI (4 seqs) & Glioma & Adult & 477-660 & Expert-revised labels, pre-operative scan bias, n=19 centres. \\
BraTS '21-23(GLI) & MRI (4 seqs) & Glioma & Adult & 2,040 & Pre-operative scan bias, multi-site diversity. \\
\citet{Pernet2016} & MRI (3 seqs) & Tumor & Adult  & 22 & Small sample size, pre-operative.\\
\citet{Cheng2017} (BTD) & MRI (T1w) & Tumor & Adult & 233 & Monocentric, private.\\
\hline

\makebox[0pt][l]{\textbf{Stroke}} & & & & & \\
ISLES 2015 (SISS) & MRI (4 seqs) & Sub-acute isch. Stroke & Adult & 64 & Mostly monocentric, ischemic stroke bias. \\
ISLES 2015 (SPES) & MRI (4 seqs) & acute isch. Stroke & Adult & 50 & Monocentric, ischemic stroke bias. \\
ATLAS v1.1/1.2 & MRI (T1w) & Chronic Stroke & Adult & 229-304 & Modality bias (T1w only), multicentric. \\
ATLAS v2.0 & MRI (T1w) & Chronic Stroke & Adult & 955 & Modality bias (T1w only), multicentric. \\
\hline

\makebox[0pt][l]{\textbf{Multiple Sclerosis (MS) and White Matter Hyperintensities (WMH)}} & & & & & \\
ISBI 2015 & MRI (4 seqs) & Multiple Sclerosis & Adult & 82  & Mono-site bias (JHU, USA). \\
MSSEG& MRI (4 seqs) & Multiple Sclerosis & Adult & 53 & Small sample size. \\
MSLUB & MRI (3 seqs) & Multiple Sclerosis & Adult & 30 & Mono-site bias (UMCL, Slovenia). \\
WMH 2017 & MRI (T1w, FLAIR) & White Matter Hyp. & Adult & 170 & Age bias (elderly), scanner bias (3 sites). \\
\hline

\makebox[0pt][l]{\textbf{Dementia}} & & & & & \\
ADNI (1/2/3) & MRI, PET & Alzheimer & Adult & $>$2,000 & Clinical trial participants (elderly). \\
OASIS-3 & MRI, PET & Alzheimer & Adult & 1,098 & Mono-site bias, longitudinal study characteristics. \\
\hline

\makebox[0pt][l]{\textbf{Healthy Controls}} & & & & & \\
UK Biobank & MRI (Various) & Healthy & Adult & $\sim$50,000 & Volunteer bias (healthier/wealthier demographics). \\
HCP (Young) & MRI (T1w, T2w) & Healthy & Young Adult & $\sim$1,200 & Age bias (22-35y), high-quality bias. \\
Cam-CAN & MRI (T1w, T2w) & Healthy & Adult & $\sim$650 & Mono-site bias (Cambridge), lifestyle homogeneity. \\
IXI & MRI (T1w, T2w, PD) & Healthy & Adult & $\sim$600 & Multi-site (3 London hospitals), demographic bias. \\
NFBS & MRI (T1w) & Healthy & Young Adult & 125 & Mono-site bias, Age bias (21-45), healthy/addict/phobic subjects.\\
Kirby & MRI (Various) & Healthy & Adult & 21 & Mono-site bias, Small sample size. \\
IBC & MRI (Various) & Healthy & Adult & 12 & Mono-site bias, Small sample size.\\

\hline

\makebox[0pt][l]{\textbf{Others}} & & & & & \\
Center-TBI & CT & Traumatic Brain Injury & Adult & $\sim$4,500 & Diverse injury mechanisms, multicentric (Europe). \\
MOOD 2020 & MRI (Brain) & Synthetic & Adult & $\sim$800 & Synthetic bias (anomalies injected into healthy data from young HCP). \\
fastMRI+ & MRI (T1w, T2w, FLAIR) & Not reported & Clinical Population  & 1,001 & Pathology/Healthy mix, axial only. \\

\end{longtable}
        \begin{longtable}{p{4.2cm}p{1.3cm}p{3.2cm}p{3cm}p{4cm}p{3.3cm}}
\caption{Comparison of unsupervised anomaly detection methods and their experimental setups.
MS: Multiple Sclerosis; TBI: Traumatic Brain Injury; WMH: White Matter Hyperintensities.
An asterisk (*) denotes the presence of a working link to a GitHub repository.  
$^\mathcal{H}$ denotes the use of healthy samples drawn from the same dataset as the pathological test set.
$^{\mathcal{P}}$ denotes the use of proxy or self-supervised objectives during training (e.g., masking, synthetic corruption). Specific proxy mechanisms are described in the corresponding architecture subsections.
Input Dimensionality refers to the sampling or patching strategy used during model training.
}
\label{tab:method_summary}
\\
\toprule
\textbf{Method} & \textbf{Arch.} & \textbf{Train Dataset} & \textbf{Test Dataset} & \textbf{Imaging Modality} & \textbf{Input Dimensionality} \\
\midrule
\endfirsthead
\toprule
\textbf{Method} & \textbf{Arch.} & \textbf{Train Dataset} & \textbf{Test Dataset} & \textbf{Imaging Modality} & \textbf{Input Dimensionality} \\
\midrule
\endhead
\midrule
\multicolumn{6}{r}{\textit{Continued on next page}} \\
\midrule
\endfoot
\bottomrule
\endlastfoot

\multicolumn{6}{l}{\textbf{Brain Tumours}} \\
\midrule
\citet{Baur2021-2} & AE & In-house &  In-house (Tumour) & FLAIR & Full 3D \\
\citet{Ghorbel2023} & AE & OASIS-3$^\mathcal{H}$ & BraTS'20 & FLAIR & Patch 2D axial \\
\citet{Kascenas2023}~*$^\mathcal{P}$ & AE & BraTS'21$^\mathcal{H}$ & BraTS'21 & T1w, T1w Gd, T2w \& FLAIR & Full 2D axial \\
 & & In-house$^\mathcal{H}$ & In-house (Multiple) & CT &  \\
\citet{Luo2023}~* & AE & In-house & BraTS'19 & T2w & Full 3D \\
\citet{Meissen2023}~* & AE & Cam-CAN\footnote{\citet{Shafto2014}} \& MOOD\footnote{\citet{Zimmerer2020}} & BraTS'20 & T1w & Partial 2D axial \\
\citet{Jimenez-Garcia2024}~$^\mathcal{P}$ & AE & IXI & BraTS'20 & T1w, T1w Gd, T2w \& FLAIR & Full 3D \\
\citet{Lu2024}~$^\mathcal{P}$ & AE & BTD~\footnote{\citet{Cheng2017}}$^\mathcal{H}$ & BTD & T1w Gd & Full 2D patch \\
\citet{Qu2026} & AE & BraTS'21$^\mathcal{H}$ & BraTS'21 & T1w, T1w Gd, T2w \& FLAIR & Full 2D axial\\
\citet{Sato2019} & VAE & IXI & BraTS'17 & T1w, T2w & Full 2D axial\\
\citet{Uzunova2019}~$^\mathcal{P}$ & VAE & BraTS'15$^\mathcal{H}$ & BraTS'15 & T1w Gd, T2w, FLAIR & Full 2D axial \& 3D \\
\citet{Zimmerer2019}~* & VAE & HCP & BraTS'17 & not reported & Full 2D axial \\
\citet{Bengs2021}~$^\mathcal{P}$ & VAE & In-house & BraTS'19 & T1w & Full 3D \\
\citet{Lambert2021} & VAE & ADNI$^\mathcal{H}$ \& IBC \& Kirby & BraTS'18 & FLAIR & Full 3D \\
\citet{Chatterjee2022}~*$^\mathcal{P}$ & VAE & IXI \& MOOD$^\mathcal{H}$ & BraTS'17 & T1w Gd, T2w & Full 2D axial \\
\citet{Pinaya2022-2} & VAE & UKB$^\mathcal{H}$ & BraTS'18 & FLAIR & Full 2D ax. \& 3D\\
\citet{Lüth2023} & VAE & HCP & BraTS'17 & T2w & Full 2D axial \\
\citet{Wijanarko2024}  & VAE & IXI & BraTS'20 & T2w & Full 2D axial \\
\citet{Dey2021} & GAN & BraTS'19$^\mathcal{H}$ & BraTS'19 & not reported & Full 2D axial \\
\citet{Nguyen2021} & GAN & NFBS & \textit{Pernet et al.}\footnote{\citet{Pernet2016}} & T1w & Full 2D axial \\
\citet{Wu2021} & GAN & BraTS'12/18$^\mathcal{H}$ & BraTS'12/18 & T2w & Patch 3D \\
\citet{Pinaya2022-1} & Diff. & UKB$^\mathcal{H}$ &  BraTS'18 & FLAIR & Full 2D axial\\
\citet{Wyatt2022}~* & Diff. & NFBS & \textit{Pernet et al.}\footnote{\citet{Pernet2016}} & T1w &  Partial 2D axial \\
\citet{Behrendt2023}~* & Diff. & IXI & BraTS'21 & T2w & Patch 2D axial \\
\citet{Iqbal2023}~*$^\mathcal{P}$ & Diff. & IXI & BraTS'21 & T2w & Full 2D axial \\
\citet{Behrendt2024}~* & Diff. & IXI & BraTS'21 & T2w & Partial 3D \\
\citet{Fontanella2024}~$^\mathcal{P}$ & Diff. & BraTS'21$^\mathcal{H}$ & BraTS'21 & T1, T1w Gd, T2w \& FLAIR & Partial 2D axial \\
\citet{Kumar2024}~*$^\mathcal{P}$ & Diff. & IXI & BraTS'20/21 & T2w & Full 2D axial \\
\citet{Bi2025}~* & Diff. & BraTS'23$^\mathcal{H}$ & BraTS'23 & FLAIR & Partial 2D axial \\
\citet{Beizaee2026}~*$^\mathcal{P}$ & Diff. & BraTS'21$^\mathcal{H}$ & BraTS'21 & T1w, T1w Gd, T2w \& FLAIR & Partial 2D axial\\
\hline

\multicolumn{6}{l}{\textbf{Stroke (Acute \& Chronic)}} \\
\midrule
\citet{Luo2023}~* & AE & In-house & In-house (Stroke)  & T2w & Full 3D \\
\citet{Qu2026} & AE & In-house & In-house (Stroke) & CT & Full 2D axial\\
\citet{Sato2019} & VAE & IXI & ATLAS 1.1 & T1w & Full 2D axial\\
\citet{Bengs2021}~$^\mathcal{P}$ & VAE & In-house & ATLAS 1.1 & T1w & Full 3D \\
\citet{Lüth2023} & VAE & HCP & ISLES 2015 & T2w & Full 2D axial \\
\citet{Bercea2023}~*$^\mathcal{P}$ & Diff. & IXI \& FastMRI+ & ATLAS 2.0 & T1w & Partial 2D axial \\
\citet{Behrendt2024}~* & Diff. & IXI & ATLAS 2.0 & T1w & Partial 3D \\
\citet{Bercea2024}~*$^\mathcal{P}$ & Diff. & IXI & ATLAS 2.0 & T1w & Full 2D axial \\
\citet{Beizaee2026}~*$^\mathcal{P}$ & Diff. & ATLAS 2.0$^\mathcal{H}$ & ATLAS 2.0 & T1w & Partial 2D axial\\
\hline

\multicolumn{6}{l}{\textbf{Multiple sclerosis (MS) \& White-matter hyperintensities (WMH)}} \\
\midrule
\citet{Baur2018} & AE & In-house & In-house (MS) & T1w, FLAIR & Full 2D axial \\
\citet{Baur2021-2} & AE & In-house &  In-house (MS) & FLAIR & Full 3D \\
\citet{Ghorbel2023} & AE & OASIS-3$^\mathcal{H}$ & MSLUB & FLAIR & Patch 2D axial \\
\citet{Luo2023}~* & AE & In-house & In-house (MS) & T2w & Full 3D \\
\citet{Lambert2021} & VAE & ADNI$^\mathcal{H}$ \& IBC \& Kirby & MSSEG, WMH & FLAIR & Full 3D \\
\citet{Pinaya2022-2} & VAE & UKB$^\mathcal{H}$ & MSLUB & FLAIR & Full 2D ax. \& 3D\\
 & & & WMH & FLAIR & \\
\citet{Dey2021} & GAN & ISBI'15$^\mathcal{H}$ & ISBI'15 & T2w or FLAIR not reported & Full 2D axial \\
\citet{Pinaya2022-1} & Diff. & UKB$^\mathcal{H}$ & MSLUB & FLAIR & Full 2D axial\\
 & & & WMH & FLAIR &  \\
\citet{Behrendt2023}~* & Diff. & IXI & MSLUB & T2w & Patch 2D axial \\
\citet{Iqbal2023}~*$^\mathcal{P}$ & Diff. & IXI & MSLUB & T2w & Full 2D axial \\
\citet{Behrendt2024}~* & Diff. & IXI & MSLUB & T2w & Partial 3D \\
 & & & WMH & T1w & \\
\citet{Fontanella2024}~$^\mathcal{P}$ & Diff. & BraTS'21$^\mathcal{H}$ & WMH & T1w, FLAIR & Partial 2D axial \\
\citet{Kumar2024}~*$^\mathcal{P}$ & Diff. & IXI & MSLUB & T2w & Full 2D axial \\
\hline

\multicolumn{6}{l}{\textbf{Traumatic Brain Injury (TBI)}} \\
\midrule
\citet{Simarro2020} & GAN & Center-TBI\footnote{\citet{Steyerberg2019}}$^\mathcal{H}$ & Center-TBI & CT & Full 3D\\
\end{longtable}
        \begin{longtable}{lcccc}
\caption{Study-level methodological appraisal of the included studies. This structured assessment summarises key methodological characteristics that may influence the interpretation and reproducibility of the included studies. \textbf{Single-centre evaluation} indicates at least one downstream evaluation relied on a single-centre dataset. \textbf{Internal evaluation} indicates that healthy training subjects were extracted from the same parent dataset used for pathological testing. \textbf{Post-hoc threshold optimisation} indicates threshold optimisation performed after model training to maximise Dice scores, rather than using a priori validation-set tuning. \textbf{Public code} indicates that the implementation code was publicly released to facilitate reproducibility.}
\label{tab:bias}
\\
\toprule
\textbf{Study} & \textbf{\shortstack{Single-centre evaluation}} & \textbf{\shortstack{Internal evaluation}} & \textbf{\shortstack{Post-hoc threshold optimisation}} & \textbf{\shortstack{Public code}} \\
\midrule
\endfirsthead
\toprule
\textbf{Study} & \textbf{\shortstack{Single-centre evaluation}} & \textbf{\shortstack{Internal evaluation}} & \textbf{\shortstack{Post-hoc threshold optimisation}} & \textbf{\shortstack{Public code}} \\
\midrule
\endhead
\midrule
\multicolumn{5}{r}{\textit{Continued on next page}} \\
\midrule
\endfoot
\bottomrule
\endlastfoot

\citet{Baur2021-2} & \checkmark & &  &  \\
\citet{Ghorbel2023} & \checkmark &  & \checkmark & \\
\citet{Kascenas2023} & \checkmark & \checkmark & \checkmark & \checkmark \\
\citet{Luo2023} & \checkmark &  &  & \checkmark \\
\citet{Meissen2023} &  &  &  & \checkmark \\
\citet{Jimenez-Garcia2024} &  &  &  &  \\
\citet{Lu2024} & \checkmark & \checkmark &  &  \\
\citet{Qu2026} & \checkmark & \checkmark &  &  \\
\citet{Sato2019} &  &  &  &  \\
\citet{Uzunova2019} &  & \checkmark &  &  \\
\citet{Zimmerer2019} &  &  &  & \checkmark \\
\citet{Bengs2021} & \checkmark &  &  &  \\
\citet{Lambert2021} &  &  &  &  \\
\citet{Chatterjee2022} &  &  &  & \checkmark \\
\citet{Pinaya2022-2} & \checkmark &  & \checkmark &  \\
\citet{Lüth2023} & \checkmark &  &  &  \\
\citet{Wijanarko2024} &  &  &  &  \\
\citet{Dey2021} & \checkmark & \checkmark &  &  \\
\citet{Nguyen2021} & \checkmark &  &  &  \\
\citet{Wu2021} &  & \checkmark &  &  \\
\citet{Pinaya2022-1} & \checkmark &  & \checkmark &  \\
\citet{Wyatt2022} & \checkmark &  &  & \checkmark \\
\citet{Behrendt2023} & \checkmark &  &  & \checkmark \\
\citet{Iqbal2023} & \checkmark &  & \checkmark & \checkmark \\
\citet{Behrendt2024} & \checkmark &  &  & \checkmark \\
\citet{Fontanella2024} &  & \checkmark &  &  \\
\citet{Kumar2024} & \checkmark &  & \checkmark & \checkmark \\
\citet{Bi2025} &  & \checkmark & \checkmark & \checkmark \\
\citet{Beizaee2026} &  & \checkmark & \checkmark & \checkmark \\
\citet{Bercea2023} &  &  & \checkmark & \checkmark \\
\citet{Bercea2024} &  &  & \checkmark & \checkmark \\
\citet{Baur2018} & \checkmark &  &  &  \\
\citet{Simarro2020} &  & \checkmark &  &  \\

\end{longtable}
        \begin{longtable}{lp{1.2cm}lccccp{4.5cm}}
\caption{Comparison of detection performance using AUROC and AUPRC.
Methods are grouped by model family (Autoencoders, VAEs, GANs, Diffusion models,
and related approaches), as defined in Table~\ref{tab:method_summary}.
Only studies explicitly reporting AUROC and/or AUPRC are included.}
\label{tab:detection_metrics}
\\
\toprule
\textbf{Method} & \textbf{Arch.} & \textbf{Test Dataset} & \textbf{AUROC} & \textbf{AUPRC} & \textbf{Eval. level} & \textbf{Notes} \\
\midrule
\endfirsthead
\toprule
\textbf{Method} & \textbf{Arch.} & \textbf{Test Dataset} & \textbf{AUROC} & \textbf{AUPRC} & \textbf{Eval. level} & \textbf{Notes} \\
\midrule
\endhead
\midrule
\multicolumn{7}{r}{\textit{Continued on next page}} \\
\midrule
\endfoot
\bottomrule
\endlastfoot

\multicolumn{7}{l}{\textbf{Brain Tumours}} \\
\midrule
\citet{Ghorbel2023} & AE & BraTS'20 & 0.780 & 0.425 & Pixel & \\
\citet{Kascenas2023}~* & AE & BraTS'21 & -- & 0.833 & Pixel & \\
 & & In-house (Multiple) & -- & 0.693 & Pixel & \\
\citet{Luo2023}~* & AE & BraTS'19 & 0.844 & 0.741 & Slice & \\
\citet{Meissen2023}~* & AE & BraTS'20 & 0.790 & -- & Pixel & \\
\citet{Jimenez-Garcia2024} & AE & BraTS'20 & 0.838 & -- & Voxel & \\
\citet{Lu2024} & AE & BTD\footnote{\citet{Cheng2017}} & 0.992 & -- & Slice &\\
\citet{Qu2026} & AE & BraTS'21 & 0.905 & -- & Slice & \\
\citet{Sato2019} & VAE & BraTS'17 & 0.582 & -- & Pixel & {\footnotesize Evaluated on T1w}\\
 & & & 0.788 & -- & Pixel & {\footnotesize Evaluated on T2}\\
\citet{Uzunova2019} & VAE & BraTS'15 & 0.950 & -- & Pixel & {\footnotesize Evaluated in 2D on T1w Gd}\\
 & & & 0.950 & -- & Voxel & {\footnotesize Evaluated in 3D on T1w Gd}\\
 & & & 0.940 & -- & Voxel & {\footnotesize Evaluated in 3D on fusion of T1w Gd, T2w \& FLAIR}\\
\citet{Zimmerer2019}~* & VAE & BraTS'17 & 0.820 & -- & Pixel & \\
\citet{Bengs2021} & VAE & BraTS'19 & -- & 0.279 & Slice & \\
\citet{Pinaya2022-2} & VAE & BraTS'18 & -- & 0.555 & Pixel & \\
\citet{Lüth2023} & VAE &  BraTS'17 & 0.942 & 0.380 & Voxel & \\
\citet{Wijanarko2024} & VAE & BraTS'20 & 0.968 & 0.462 & Voxel & \\
\citet{Wyatt2022}~* & Diff. & \textit{Pernet et al.}\footnote{\citet{Pernet2016}}  & 0.863 & -- & Pixel & \\
\citet{Behrendt2023}~* & Diff. & BraTS'21 & -- & 0.541 & Voxel & \\
\citet{Iqbal2023}~* & Diff. & BraTS'21 & -- & 0.590 & Pixel & \\
\citet{Kumar2024}~* & Diff. & BraTS'20 & -- & 0.417 & Pixel & \\
 & & BraTS'21 & -- & 0.578 & Pixel & \\
\citet{Bi2025}~* & Diff. & BraTS'23 & 0.922 & -- & Slice & \\
\hline

\multicolumn{7}{l}{\textbf{Stroke (Acute \& Chronic)}} \\
\midrule
\citet{Luo2023}~* & AE & In-house & 0.807 & 0.705 & Slice & Evaluated on Stroke\\
\citet{Qu2026} & AE & In-house & 0.839 & -- & Slice & Evaluated on Stroke\\
\citet{Sato2019} & VAE & ATLAS 1.1 & 0.672 & -- & Pixel & \\
\citet{Bengs2021} & VAE & ATLAS 1.1 & -- & 0.256 & Slice & \\
\citet{Lüth2023} & VAE & ISLES 2015 & 0.898 & 0.186 & Voxel & \\
\citet{Bercea2023}~* & Diff. & ATLAS 2.0 & -- & 0.145 & Pixel & \\
\hline

\multicolumn{7}{l}{\textbf{Multiple Sclerosis (MS) \& White Matter Hyperintensities (WMH)}} \\
\midrule
\citet{Baur2021-2} & AE & In-house & -- & 0.300 & Voxel & Evaluated on MS \\
\citet{Ghorbel2023} & AE & MSLUB & 0.886 & 0.203 & Pixel & \\
\citet{Luo2023}~* & AE & In-house & 0.858 & 0.731 & Slice & Evaluated on MS\\
\citet{Pinaya2022-2} & VAE & MSLUB & -- & 0.272 & Pixel & \\
 & & MSLUB & 0.866 & -- & Slice & \\
 & & WMH & -- & 0.320 & Pixel & \\
\citet{Behrendt2023}~* & Diff. & MSLUB & -- & 0.106 & Voxel & \\
\citet{Iqbal2023}~* & Diff. & MSLUB & -- & 0.106 & Pixel & \\
\citet{Kumar2024}~* & Diff. & MSLUB & -- & 0.067 & Pixel & \\
\hline

\multicolumn{7}{l}{\textbf{Traumatic Brain Injury (TBI)}} \\
\midrule
\citet{Simarro2020} & GAN & Center-TBI & 0.749 & -- & Voxel & \\
\end{longtable}
        \begin{longtable}{lp{1.2cm}lclp{5.5cm}}
\caption{Comparison of segmentation performance using the Dice similarity coefficient (Dice). Methods are grouped by model family (Autoencoders, VAEs, GANs, and Diffusion models), as defined in Table~\ref{tab:method_summary}. This table details the reported Dice scores alongside the thresholding strategies employed to binarise residual maps (e.g., post-hoc optimisation, validation set, or percentile-based thresholding). Only studies reporting segmentation performance using the Dice similarity coefficient are included.}
\label{tab:dice_summary}
\\
\toprule
\textbf{Method} & \textbf{Arch.} & \textbf{Test Dataset} & \textbf{Dice} & \textbf{Thresh. Strategy} & \textbf{Notes} \\
\midrule
\endfirsthead
\toprule
\textbf{Method} & \textbf{Arch.} & \textbf{Test Dataset} & \textbf{Dice} & \textbf{Thresh. Strategy} & \textbf{Notes} \\
\midrule
\endhead
\midrule
\multicolumn{6}{r}{\textit{Continued on next page}} \\
\midrule
\endfoot
\bottomrule
\endlastfoot

\multicolumn{6}{l}{\textbf{Brain Tumours}} \\
\midrule
\citet{Baur2021-2} & AE & In-house & 0.390 & Validation set & Evaluated on Tumour\\
\citet{Luo2023}~* & AE & BraTS'19 & 0.462 & Validation set & \\
\citet{Meissen2023}~* & AE & BraTS'20 & 0.400 & Percentile-based & \\
\citet{Jimenez-Garcia2024} & AE & BraTS'20 & 0.471 & Validation set & \\
\citet{Uzunova2019} & VAE & BraTS'15 & 0.550 & Validation set & {Evaluated in 2D on T1w Gd}\\
 & & & 0.320 &  & {Evaluated in 3D on T1w Gd}\\
 & & & 0.500 &  & {Evaluated in 3D on fusion of T1w Gd, T2w \& FLAIR}\\
\citet{Zimmerer2019}~* & VAE & BraTS'17 & 0.440 & Validation set & \\
\citet{Bengs2021} & VAE & BraTS'19 & 0.301 & Validation set & \\
\citet{Lambert2021} & VAE & BraTS'18 & 0.650 & Validation set & \\
\citet{Chatterjee2022}~* & VAE & BraTS'17 & 0.531 & Histogram-based & {Evaluated on T1w Gd}\\
 & & & 0.642 & & {Evaluated on T2w}\\
\citet{Wijanarko2024} & VAE & BraTS'20 & 0.606 & Fixed & \\
\citet{Dey2021} & GAN & BraTS'19 & 0.680 & Histogram-based & \\
\citet{Nguyen2021} & GAN & \textit{Pernet et al.}\footnote{\citet{Pernet2016}} & 0.770 & Histogram-based & \\
\citet{Wu2021} & GAN & BraTS'12/18 & 0.630 & Percentile-based & \\
\citet{Wyatt2022}~* & Diff. & \textit{Pernet et al.}\footnote{\citet{Pernet2016}} & 0.383 & Fixed & \\
\citet{Behrendt2023}~* & Diff. & BraTS'21 & 0.490 & Validation set & \\
\citet{Behrendt2024}~* & Diff. & BraTS'21 & 0.574 & Validation set & \\
\citet{Fontanella2024} & Diff. & BraTS'21 & 0.699 & Validation set & \\
\addlinespace
\multicolumn{6}{l}{\textit{\small --- Post-hoc test-set optimisation ---}} \\
\addlinespace
\citet{Ghorbel2023} & AE & BraTS'20 & 0.502 & & \\
\citet{Kascenas2023}~* & AE & BraTS'21 & 0.773 & & \\
 & & In-house (Multiple) & 0.674 & & \\
\citet{Pinaya2022-2} & VAE & BraTS'18 & 0.537 & & {Evaluated in 2D}\\
 & & BraTS'18 & 0.617 & & {Evaluated in 3D}\\
\citet{Pinaya2022-1} & Diff. & BraTS'18 & 0.398 & & \\
\citet{Iqbal2023}~* & Diff. & BraTS'21 & 0.530 & & \\
\citet{Kumar2024}~* & Diff. & BraTS'20 & 0.371 & & \\
 & & BraTS'21 & 0.506 &  & \\
\citet{Bi2025}~* & Diff. & BraTS'23 & 0.738 & & \\
\citet{Beizaee2026}~* & Diff. & BraTS'21 & 0.697 & & {Evaluated on T1w}\\
 & & & 0.730 & & {Evaluated on T1w Gd}\\
 & & & 0.803 & & {Evaluated on T2w}\\
 & & & 0.851 & & {Evaluated on FLAIR}\\
\hline

\multicolumn{6}{l}{\textbf{Stroke (Acute \& Chronic)}} \\
\midrule
\citet{Bengs2021} & VAE & ATLAS 1.1 & 0.331 & Validation set & \\
\citet{Behrendt2024}~* & Diff. & ATLAS 2.0 & 0.148 & Validation set & \\
\addlinespace
\multicolumn{6}{l}{\textit{\small --- Post-hoc test-set optimisation ---}} \\
\addlinespace
\citet{Bercea2023}~* & Diff. & ATLAS 2.0 & 0.228 & & \\
\citet{Bercea2024}~* & Diff. & ATLAS 2.0 & 0.297 & & \\
\citet{Beizaee2026}~* & Diff. & ATLAS 2.0 & 0.416 & & \\
\hline

\multicolumn{6}{l}{\textbf{Multiple Sclerosis (MS) \& White Matter Hyperintensities (WMH)}} \\
\midrule
\citet{Baur2018} & AE & In-house & 0.605 & Percentile-based & Evaluated on MS\\
\citet{Lambert2021} & VAE & WMH, MSSEG & 0.463 & Validation set & \\
\citet{Dey2021} & GAN & ISBI'15 & 0.482 & Histogram-based & \\
\citet{Behrendt2023}~* & Diff. & MSLUB & 0.105 & Validation set & \\
\citet{Behrendt2024}~* & Diff. & MSLUB & 0.061 & Validation set & \\
 & & WMH & 0.132 & Validation set & \\
\citet{Fontanella2024} & Diff. & WMH & 0.569 & Validation set & \\
\addlinespace
\multicolumn{6}{l}{\textit{\small --- Post-hoc test-set optimisation ---}} \\
\addlinespace
\citet{Ghorbel2023} & AE & MSLUB & 0.173 & & \\
\citet{Pinaya2022-2} & VAE & MSLUB & 0.378 & & {Evaluated in 2D}\\
 & & WMH & 0.429 & & {Evaluated in 2D}\\
 & & MSLUB & 0.133 & & {Evaluated in 3D}\\
 & & WMH & 0.133 & & {Evaluated in 3D}\\
\citet{Pinaya2022-1} & Diff. & MSLUB & 0.247 & & \\
 & &  WMH & 0.298 & & \\
\citet{Iqbal2023}~* & Diff. & MSLUB & 0.107 & & \\
\citet{Kumar2024}~* & Diff. & MSLUB & 0.055 & & \\
\end{longtable}
\end{landscape}


\subsection{Methodological evolution}
\subsubsection{Autoencoders}
\paragraph{}
Autoencoders do not define an explicit probabilistic density over the data but instead learn a deterministic encoder-decoder mapping that reconstructs input images through a compressed latent representation. In unsupervised anomaly detection, AEs are trained on healthy data to model normative anatomical variability; anomalies are inferred from residual differences between the input and its reconstruction. Although not strictly generative in a probabilistic sense, AEs established the reconstruction-based paradigm that underpins later probabilistic approaches such as variational autoencoders and diffusion models.
\paragraph{Architecture recap.}
Autoencoders were among the first deep learning architectures applied to unsupervised anomaly detection in neuroimaging. They consist of an encoder–decoder structure trained to reconstruct inputs through a compressed latent representation. In UAD settings, AEs are trained on healthy data, and anomalies are inferred from residual differences between input and reconstruction. Early implementations relied on fully connected bottlenecks, which imposed strong compression and limited spatial correspondence between latent representations and anatomical structures, often resulting in oversmoothed reconstructions~\citep{Baur2018}. Subsequent work adopted fully convolutional architectures with skip connections and residual blocks to better preserve spatial information. Residual connections were introduced to facilitate gradient flow in deeper networks \citep{He2016}, and some studies additionally incorporated attention mechanisms inspired by vision transformers~\citep{Dosovitskiy2020, Ghorbel2023, Qu2026}.
\paragraph{Architectural variations.}
Several extensions were introduced within the AE framework. Denoising autoencoders treated pathological regions as structured corruption and reconstructed plausible healthy patterns \citep{Kascenas2023}. Training strategies included elastic deformation-based augmentations \citep{Jimenez-Garcia2024} and composite reconstruction losses integrating contrastive or discriminative components \citep{Lu2024}. Some approaches incorporated pretrained convolutional feature extractors to guide latent representations \citep{Meissen2023}. Both 2D slice-wise and full 3D volumetric implementations were reported; volumetric models increased spatial context but required careful tuning of latent dimensionality to avoid oversmoothing or partial reconstruction of anomalous structures \citep{Luo2023}.

\subsubsection{Variational Autoencoders (VAE)}
\paragraph{Architecture recap.}
Variational autoencoders extend the deterministic autoencoder framework by introducing a probabilistic latent space. Instead of mapping each input to a single latent code, the encoder outputs the parameters of a distribution - typically a Gaussian defined by mean and variance -regularised toward a prior, most often $\mathcal{N}(0, I)$ \citep{Kingma2013}. During training, latent samples are drawn from this parameterised posterior distribution and passed through the decoder to reconstruct the input. The learning objective combines a reconstruction term with a Kullback–Leibler (KL) divergence that encourages alignment between the learned latent distribution and the prior. In unsupervised anomaly detection, anomalies are inferred from reconstruction residuals and/or latent-space deviations.

\paragraph{Architectural variations.}
Several extensions of the canonical VAE were proposed. Spatial VAEs preserved the latent representation as a low-resolution feature map rather than collapsing it into a dense vector, maintaining spatial correspondence between latent units and anatomical regions \citep{Lambert2021, Bengs2021}. Vector-quantised VAEs (VQ-VAEs) replaced continuous latent variables with discrete codebooks of embeddings to capture more structured latent representations \citep{Oord2017}. Building on this formulation, transformer-based autoregressive models were trained on healthy data to modify pathological latent codes toward healthy counterparts \citep{Marimont2021, Pinaya2022-2}.

Loss formulations also evolved. Some studies modified reconstruction terms or latent regularisation to alter sensitivity to anatomical variation \citep{Sato2019}. Structural similarity measures and composite reconstruction objectives integrating $\ell_1$, $\ell_2$, and perceptual terms were explored \citep{Meissen2023, Wijanarko2024}. Context-encoding VAEs (ceVAEs) combined probabilistic reconstruction with masked inpainting branches sharing encoder–decoder weights to encourage contextual feature learning \citep{Chatterjee2022}. Additional variations included contrastive pretraining stages and alternative decoder distributions such as Gaussian mixture models \citep{Koller2009} and normalising flows \citep{Lüth2023, Rezende2015}. 

Conditioning mechanisms and positional encodings were also investigated in both 2D and 3D settings \citep{Uzunova2019}. Anomaly scoring strategies ranged from pure reconstruction residuals to ELBO-informed metrics incorporating the KL divergence term \citep{Zimmerer2019}.

\subsubsection{Generative Adversarial Networks (GANs)}
\paragraph{Architecture recap.}
Generative adversarial networks, introduced by \citet{Goodfellow2014}, consist of a generator trained to synthesise samples and a discriminator trained to distinguish generated from real data. Through adversarial optimisation, the generator learns to produce samples that approximate the distribution of the training data. In neuroimaging UAD, GANs are typically trained on healthy brain images and used to generate pseudo-healthy reconstructions of pathological inputs, with anomalies inferred from residual differences between input and reconstruction. 
The most commonly adopted medical UAD framework was f-AnoGAN \citep{Schlegl2019}, an extension of AnoGAN \citep{Schlegl2017} incorporating an encoder for faster inference and a Wasserstein-based objective to improve training stability. As with other adversarial models, GANs remain sensitive to training instability and mode collapse.

\paragraph{Architectural variations.}
Several task-specific adaptations were introduced within the GAN framework. Symmetry-driven designs incorporated contralateral hemisphere information as a structural prior for anomaly localisation \citep{Wu2021}. Partition-based strategies separated candidate anomalous regions prior to adversarial evaluation \citep{Dey2021}. Multi-stage pipelines introduced additional refinement modules to improve coarse reconstructions \citep{Nguyen2021}. 
Both 2D slice-wise and 3D implementations were represented among the included GAN studies (Table~\ref{tab:method_summary}).

\subsubsection{Diffusion models}
\paragraph{Architecture recap.}
Most diffusion-based UAD methods rely on Denoising Diffusion Probabilistic Models (DDPMs) \citep{Ho2020}. In DDPMs, a forward process progressively corrupts training images by adding Gaussian noise over multiple timesteps. A neural network - typically a U-Net \citep{Ronneberger2015} - is trained to predict the added noise, thereby learning to reverse the corruption process.
In neuroimaging UAD, diffusion models are typically trained on healthy images, such that the learned denoising process implicitly models normal anatomical variability. During inference, pathological inputs are partially noised and subsequently denoised toward this learned healthy manifold. Anomaly maps are derived from discrepancies between the input and the resulting pseudo-healthy reconstruction \citep{Pinaya2022-1, Wyatt2022, Bercea2023}.
Because voxel-space DDPMs are computationally demanding for high-resolution 3D MRI, several studies adopted latent diffusion strategies in which images are first compressed via an autoencoder and diffusion is performed in latent space \citep{Pinaya2022-1, Kumar2024}. This reduces memory and computational requirements while preserving structural content.

\paragraph{Architectural variations.}
Several methodological directions were explored within diffusion-based UAD.

\textit{Noise design.} While early approaches used Gaussian corruption, structured noise patterns such as Perlin or Simplex noise were investigated to modify the corruption process \citep{Wyatt2022, Bercea2024}.

\textit{Similarity metrics.} Beyond voxel-wise intensity residuals, structure-aware metrics such as SSIM \citep{Behrendt2024} and perceptual feature-based measures including LPIPS \citep{Zhang2018, Chen2019} were used to define anomaly maps.

\textit{Guided restoration.} To mitigate excessive modification of healthy tissue during denoising, some methods reintroduced uncorrupted image regions after an initial pass \citep{Bercea2023} or selectively preserved anatomical regions during reconstruction \citep{Bercea2024}, while others, like MCDDPM~\citep{Kumar2024}, utilise multichannel latent representations and cross-attention to integrate contextual information directly into the denoising process.

\textit{Patch- and mask-based designs.} Patch-based diffusion models restricted corruption to selected spatial regions \citep{Behrendt2023}, while masked DDPM variants modified spatial or frequency components of the input \citep{Iqbal2023}.

\textit{Hybrid and multi-stage inference.} Some pipelines combined diffusion with saliency-driven or counterfactual guidance mechanisms \citep{Fontanella2024}. Multi-stage inference cascades applied repeated denoising passes to progressively attenuate pathological structures \citep{Bi2025}.

\textit{Emerging continuous-time frameworks.} Beyond stochastic DDPMs, deterministic alternatives based on conditional flow matching and rectified flows were introduced \citep{Lipman2023, Albergo2023, Liu2023}. Early neuroimaging implementations such as REFLECT \citep{Beizaee2026} operated in latent space and relied on self-supervised proxy corruptions to construct training objectives.
\paragraph{} While architectures evolved to generate increasingly realistic pseudo-healthy reconstructions, their reported anomaly-localisation performance nevertheless varies substantially across pathology types and lesion morphologies. We therefore synthesise the evidence by pathology while considering architectural family within each pathology group.

\subsection{Pathology-stratified synthesis}
\subsubsection{Brain Tumours (large / focal lesions)}
\paragraph{Study characteristics.}
Brain tumours, particularly gliomas represented in the BraTS datasets, constitute the most extensively studied pathology in the UAD literature~\citep{Menze2015}. These lesions are characterised by large focal volumes, substantial mass effect, and high contrast against healthy tissue on structural sequences~\citep{Menze2015}. Models across all four architectural families (AE, VAE, GAN, and Diffusion) have been evaluated on this pathology. The most commonly used MRI sequences were T1-w Gd, T2-w, and FLAIR, with one study also incorporating CT~\citep{Kascenas2023}. Both 2D slice-wise and full 3D volumetric approaches were widely employed, with neither approach demonstrating a consistent advantage across studies.
\paragraph{Quantitative synthesis.}
Segmentation performance for tumours spanned a wide interval. When using pre-specified or validation-set thresholding, reported Dice scores ranged from 0.301 \citep{Bengs2021} to 0.770 \citep{Nguyen2021}. However, when employing retrospective, post-hoc threshold optimisation, Dice scores reached up to 0.851 \citep{Beizaee2026}. Detection metrics similarly showed high peak values: voxel/pixel-level AUROC ranged from 0.582 \citep{Sato2019} to 0.968 \citep{Wijanarko2024}, while slice-level AUROC peaked at 0.992 \citep{Lu2024}. Voxel/pixel-level AUPRC ranged from 0.380 \citep{Lüth2023} to 0.833 \citep{Kascenas2023}.
\paragraph{Key findings.}
Despite substantial methodological evolution across architectural families, tumour segmentation scores overlap considerably across studies. No consistent improvement in reported performance is apparent across increasingly complex architectural families, although heterogeneous experimental conditions preclude direct attribution to architectural design. One possible explanation for this overlap is the reconstruction paradox: highly expressive models may learn identity mappings, faithfully reconstructing anomalous tumour tissue rather than removing it, thereby reducing the residual contrast required for reliable anomaly localisation. Furthermore, post-hoc threshold optimisation yields optimistic upper-bound Dice estimates that may not reflect clinically deployable operating points.

\subsubsection{Stroke (acute \& chronic)}
\paragraph{Study characteristics.}
Stroke lesions exhibit substantial morphological heterogeneity, ranging from acute infarcts to chronic tissue loss and structural deformation. Stroke was addressed by AE, VAE, and diffusion architectures, but no GAN-based UAD studies met the inclusion criteria for this pathology. Most studies evaluated T1-w images from the ATLAS datasets, alongside T2-w and CT from ISLES 2015 and institutional cohorts.
\paragraph{Quantitative synthesis.}
Reported segmentation performance for stroke was generally lower than that reported for brain tumours. Using validation-set thresholding, Dice scores ranged from 0.148 \citep{Behrendt2024} to 0.331 \citep{Bengs2021}. With post-hoc optimised thresholds, Dice reached up to 0.416 \citep{Beizaee2026}. Voxel/pixel-level AUROC ranged from 0.672 \citep{Sato2019} to 0.898 \citep{Lüth2023}, while slice-level AUROC reached 0.839 \citep{Qu2026}. Voxel-level AUPRC values ranged from 0.145 \citep{Bercea2023} to 0.256 \citep{Bengs2021}.
\paragraph{Key findings.}
The lower reported performance for stroke highlights a potential limitation of reconstruction-based UAD for heterogeneous lesions and chronic tissue loss. Such abnormalities may be particularly challenging when lesion appearance overlaps with cerebrospinal fluid or lacks sharp intensity gradients. The reliance on single-sequence structural imaging (e.g., T1-w in ATLAS) may further limit the residual contrast, suggesting that multi-parametric representations could potentially improve the distinction between chronic infarcts and normative anatomical variability.

\subsubsection{Multiple Sclerosis (MS) \& White Matter Hyperintensities (WMH) (small / diffuse lesions)}
\paragraph{Study characteristics.}
Multiple Sclerosis plaques and WMH are characterised by small, sparse, and spatially distributed lesions, predominantly visualised as hyperintensities on FLAIR and T2-w sequences. Because these lesions occupy only a tiny fraction of the total brain volume, they introduce severe class imbalance into the evaluation. Studies evaluating these lesions utilised datasets such as MSLUB, MSSEG, ISBI'15, and WMH 2017, applying models across all four architectural families in both 2D and 3D configurations.
\paragraph{Quantitative synthesis.}
Reported segmentation performance was generally lower for these small and diffuse lesions. Several MS studies reported Dice scores below 0.30, with diffusion and VAE studies reporting values ranging from 0.055 \citep{Kumar2024} to 0.482 \citep{Dey2021}. An early AE study \citep{Baur2018} reported a Dice of 0.605, illustrating the variability across studies. For WMH, reported Dice scores ranged from 0.132 \citep{Behrendt2024} to 0.569 \citep{Fontanella2024}. While voxel/pixel-level AUROC reached as high as 0.886 \citep{Ghorbel2023}, voxel/pixel-level AUPRC remained low, ranging from 0.067 \citep{Kumar2024} to 0.300 \citep{Baur2021-2} for MS and reaching 0.320 \citep{Pinaya2022-2} for WMH.
\paragraph{Key findings.}
Current generative UAD methods do not yet provide reliable out-of-the-box segmentation for small or diffuse lesions. Crucially, evaluating these lesions exposes a distinct inconsistency in UAD metrics: several studies report high voxel-level AUROC alongside poor Dice and AUPRC scores. As an illustrative example, \citet{Ghorbel2023} reported a pixel-level AUROC of 0.886 on the MSLUB dataset, compared with an AUPRC of 0.203 and a Dice score of 0.173 under post-hoc thresholding optimisation. In the presence of extreme class imbalance, voxel/pixel-level AUROC should therefore be interpreted with caution because it may remain high despite limited lesion localisation performance. AUPRC and Dice provide complementary information to AUROC when evaluating sparse lesions, although Dice remains sensitive to threshold selection and lesion size. Notably, none of the included studies reported lesion-wise evaluation metrics, making it impossible to distinguish whether low segmentation performance reflects failure of lesion detection or imprecise lesion delineation.

\subsubsection{Traumatic Brain Injury (TBI)}
\paragraph{Study characteristics \& synthesis}
Traumatic Brain Injury was addressed by a single included study using a 3D GAN architecture on CT imaging from the CENTER-TBI cohort \citep{Simarro2020}. The study reported a voxel-level AUROC of 0.749 for localising acute TBI abnormalities. No Dice or AUPRC metrics were reported for this task. Given the limited representation of TBI in the generative UAD literature, a robust comparative analysis of architectural differences for this pathology is not yet feasible.

\subsection{Cross-study synthesis}
\begin{figure}[]
    \centering
    \includegraphics[width=\linewidth]{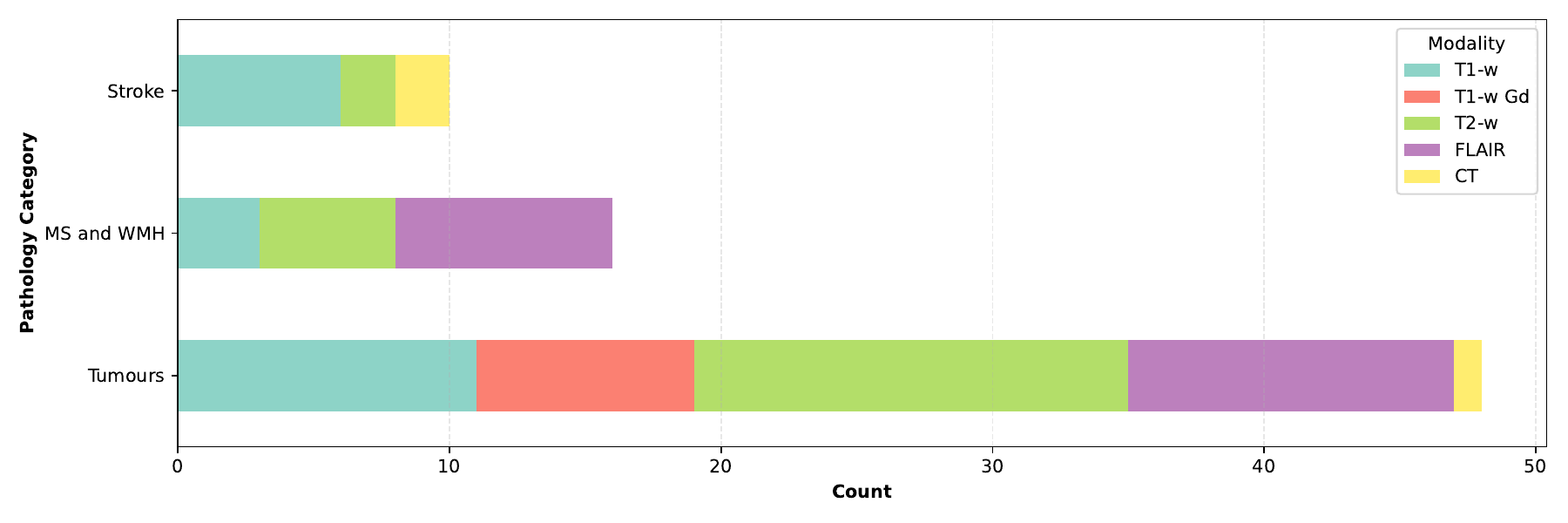}
    \caption{Distribution of MRI sequences and imaging modalities across major pathology categories.}
    \label{fig:modality_stackedbars}
\end{figure}

Unsupervised brain tumour segmentation has been investigated more extensively than other pathologies, likely reflecting the relatively high availability of curated datasets such as BraTS. MRI remained the predominant imaging modality across all pathologies, with T1-w, T2-w, and FLAIR being the most frequently utilised sequences (Figure~\ref{fig:modality_stackedbars}). T1-w Gd was commonly used in glioma datasets, while CT was sporadically employed for stroke or traumatic brain injury.

Input dimensionality varied considerably. In total, 10 studies adopted full volumetric 3D processing, whereas 16 processed complete 2D axial stacks sequentially. A further 7 studies used partial 2D processing restricted to selected axial slices, typically centred around mid-brain regions. Overall, 2D processing remained predominant, particularly among diffusion-based approaches.

Dice thresholding standardisation represents a major obstacle for cross-study comparison. Ten studies reported an upper-bound metric obtained through retrospective selection of the threshold yielding the best possible Dice, whereas 10 employed validation-set thresholding. The remaining eight used fixed or algorithmic thresholding strategies (e.g., Otsu, histogram-based, Felzenszwalb). Reporting of post-hoc optimised Dice scores became more frequent after 2022, complicating temporal comparisons and potentially inflating reported segmentation performance.

Across the reviewed studies, smaller and sparser lesions, particularly MS and WMH, were generally associated with lower reported Dice values than large focal tumours. Sub-acute and chronic stroke segmentation was less frequently studied (six studies), and reported scores were also generally lower than those reported for tumours.

\section{Discussion}
\paragraph{Principal findings.}
This scoping review synthesised eight years (2018-2025) of research on unsupervised deep generative models for anomaly detection in structural neuroimaging. Across the reviewed studies, reported performance varied substantially across pathology types and lesion morphologies, with large focal lesions generally associated with higher reported metrics than small, sparse, or heterogeneous abnormalities. However, substantial experimental heterogeneity prevents the relative contributions of pathology-related factors and architectural design from being reliably disentangled.

Brain tumours - typically large and focal - generally showed the highest reported segmentation and detection metrics. In contrast, smaller, sparse, or heterogeneous abnormalities such as multiple sclerosis (MS), white-matter hyperintensities (WMH), and stroke were generally associated with lower reported performance, with Dice scores frequently falling below 0.50, and sometimes below 0.30. Ultimately, no unsupervised family approached the performance of contemporary supervised baselines on curated benchmarks such as BraTS~\citep{Bonato2025}, where Dice scores above 0.90 are now routinely reported~\footnote{See \url{https://www.synapse.org/Synapse:syn53708249}}. 

Despite these limitations, reconstruction-based generative models offer a distinct conceptual advantage: the ability to produce subject-specific pseudo-healthy images that explicitly visualise deviation from normative anatomy, resembling the clinical image interpretation process.

\paragraph{Results interpretation.}
The observed cross-study performance patterns may reflect interactions between lesion characteristics and model behaviour. Large, well-contrasted tumours may provide favourable conditions for residual-based detection because of their volumetric prominence and mass effect. Conversely, MS and WMH lesions create severe class imbalance, making Dice highly sensitive to localisation errors. Stroke introduces additional complexity through heterogeneous intensity profiles and ill-defined boundaries~\citep{HernandezPetzsche2022}, which may reduce residual conspicuity even when lesion volumes are large.

Architectural refinements have included expressive diffusion models \citep{Ho2020}, 3D volumetric processing \citep{Uzunova2019, Pinaya2022-2}, and attention-based masking. However, reported performance ranges remain overlapping across architectural families, and the available evidence does not allow the contributions of model design to be separated from differences in pathology, datasets, sequence selection, and evaluation protocols. Because architectures were rarely evaluated under matched conditions, the current literature provides descriptive cross-study patterns rather than definitive comparative rankings. Furthermore, reconstruction-based models remain highly sensitive to the integrity of the healthy training distribution, where demographic biases \citep{Littlejohns2020, Gianattasio2021} or even 3\% pathological contamination \citep{Behrendt2022-ex} can significantly degrade anomaly detection.

\paragraph{Methodological \& evaluation constraints of reconstruction-based UAD.}
A central limitation of most reviewed approaches lies in the reconstruction-based anomaly paradigm itself, which is vulnerable to two core scientific challenges: the reconstruction paradox and normative distribution bias. 

\textbf{The Reconstruction Paradox:} These models assume that pathological patterns are poorly represented in the learned healthy distribution and will therefore be attenuated during reconstruction. However, when lesion intensities overlap with normal anatomical variation, or when the network learns an identity shortcut (often referred to as the reconstruction paradox), anomalous structures may be partially or fully reconstructed, leading to false negatives or decreased segmentation quality. Recent work has shown that improving reconstruction fidelity alone does not guarantee improved anomaly detection performance~\citep{Meissen2022}. In such cases, voxel-wise residual maps become insufficiently discriminative, particularly for low-contrast, diffuse, or texture-preserving abnormalities.

Several mitigation strategies have been proposed. Latent-space masking techniques aim to prevent trivial identity mappings by forcing inpainting from normative representations~\citep{Beizaee2025}. Decoder-level perturbation and attention-based masking mechanisms similarly constrain shortcut learning~\citep{Tang2025}. Other approaches improve residual contrast through structured post-processing~\citep{Munoz-Ramirez2021}. Nevertheless, these solutions expose a broader trade-off: forcing modifications to prevent shortcut learning can inadvertently alter healthy tissue, while increasing reconstruction fidelity can reduce anomaly sensitivity when pathological patterns are also reconstructed, thereby diminishing residual-based contrast and increasing false negatives.

\textbf{Normative Distribution Bias:} Furthermore, generative models are sensitive to the heterogeneity of the healthy training distribution. Variations such as patient age spans, scanner differences, or even minor (e.g., 3\%) pathological contamination~\citep{Behrendt2022-ex} can severely degrade anomaly detection by shifting the model's baseline. This bias manifests as an increased number of false positives in healthy tissue. To address this, emerging works are exploring conditional generative models that explicitly condition the reconstruction on metadata to disentangle normal anatomical variance from true anomalies. Additionally, training with self-supervised synthetic corruptions~\citep{Beizaee2025} may improve robustness by encouraging models to distinguish imposed structural deviations from normative anatomy.

\paragraph{Future directions.}
Our synthesis highlights several priorities for advancing unsupervised anomaly detection (UAD) in neuroimaging. The most persistent challenge remains performance on small, sparse, or ill-defined abnormalities, for which reported Dice scores are generally lower than those reported for large focal tumours. Addressing this gap will require progress in both architectural design and evaluation methodology.

\textit{Architectural advances.}
More efficient continuous-time generative models - such as flow matching~\citep{Lipman2023} or score-based diffusion variants~\citep{Song2020-2} - offer promising alternatives to classical DDPMs, with improved sampling efficiency and potentially greater stability. Latent-space hybrids combining strong encoders (e.g., VQ-VAEs or spatial VAEs) with diffusion- or flow-based decoders may enable scalable volumetric pseudo-healthy reconstruction~\citep{Pinaya2022-2, Lambert2021, Bengs2021}. Incorporating anatomy-aware priors, including symmetry constraints or atlas guidance, may improve localisation of subtle or unilateral abnormalities. Perceptually informed residual measures (e.g., SSIM, LPIPS) and hybrid generative–discriminative strategies may further enhance anomaly contrast while limiting false positives~\citep{Behrendt2024, Zhang2018, Chen2019, Huijben2024-ex}. Nevertheless, architectural refinement alone is unlikely to address the full range of challenges associated with heterogeneous lesion characteristics and clinical imaging conditions.

\textit{Self-supervised and anomaly-aware paradigms.}
A subset of recent studies adopts self-supervised strategies in which models are exposed to synthetic anomalies during training (e.g., corrupted latent representations in \citet{Beizaee2026}). These approaches often report higher Dice scores by explicitly learning deviation patterns. However, synthetic perturbations may not faithfully capture the appearance of real-world clinical anomalies, particularly low-contrast or diffuse lesions. This introduces a conceptual distinction between purely normative generative models and anomaly-aware self-supervised frameworks, with uncertain implications for generalisability across heterogeneous pathologies.

\paragraph{Related foundation-model-based anomaly pipelines (outside normative modelling).}
Recent work has also explored foundation-model-based anomaly detection pipelines that leverage large pretrained encoders or segmentation models. For example, \citet{Ma2025} used CLIP-generated pseudo-labels and a foundation segmentation model to train a downstream 3D U-Net in a self-supervised manner. Similarly, \citet{Rahmaniar2026} proposed a teacher–student distillation framework built on an ImageNet-pretrained backbone for tumour detection. Unlike normative generative models, these approaches rely on pretrained semantic representations or pseudo-supervision and do not learn a healthy anatomical distribution through reconstruction. They were therefore considered adjacent but methodologically distinct from reconstruction-based generative UAD. Furthermore, brain imaging foundation models present a promising avenue for the few-shot segmentation of anomalies. Recently, \citet{Tak2026} demonstrated that the learned encoders of these foundation models can be readily adapted to segmentation tasks, achieving Dice scores comparable to existing UAD methods on brain tumours. However, the capacity of such foundation models to segment rare and unlabelled anomalies remains an open area for future evaluation.

\paragraph{Evaluation reform and benchmarking.}
A major bottleneck for comparative analysis remains the absence of standardised evaluation protocols. Metrics are inconsistently reported across voxel-, slice-, and subject-level tasks, which are not directly comparable. Furthermore, post-hoc threshold optimisation frequently yields upper-bound 'best achievable' Dice scores that may overestimate clinically realistic performance relative to validation-based thresholding. Notably, 10 out of the 28 studies reporting segmentation metrics relied on this retrospective thresholding strategy (Table~\ref{tab:bias}).

Beyond metric reporting, the evaluation of UAD models is constrained by dataset selection and evaluation methodology. As structured in Table~\ref{tab:method_summary} and Table~\ref{tab:bias}, 55\% (18/33) of the reviewed literature relied on strictly monocentric or institutional in-house datasets. Furthermore, a third of the studies (11/33) lacked true external validation, opting instead to extract healthy training subjects from the same dataset used for pathological testing. This reliance on internal splits and single-scanner cohorts hides potential generalisation failures and severely limits the assessment of model robustness against real-world domain shifts and demographic variability.

Collectively, these sources of heterogeneity make it difficult to determine whether reported performance differences reflect model architecture or differences in datasets and evaluation protocols, underscoring the need for standardised benchmarking under matched experimental conditions.

Conventional metrics such as Dice and AUROC assess localisation accuracy on predefined lesions but do not capture the core objective of reconstruction-based generative models: robust normative representation learning. Importantly, reconstruction fidelity is not equivalent to anomaly sensitivity; improving reconstruction quality does not guarantee improved detection, particularly when pathological patterns overlap with normal anatomical variation~\citep{Meissen2022}. Conversely, competitive Dice scores do not ensure preservation of healthy tissue or plausible correction of anomalies.

To address this gap, \citet{Bercea2025} reframed UAD as a problem of \textit{normative representation learning} and proposed task-specific indices tailored to pseudo-healthy reconstructions:
\begin{itemize}
    \item \textit{Restoration Quality Index (RQI)} --- evaluates perceptual reconstruction behaviour using LPIPS~\citep{Zhang2018}.
    \item \textit{Anomaly-to-Healthy Index (AHI)} --- measures how plausibly a pathological image is transformed toward a healthy distribution using Fréchet Inception Distance.
    \item \textit{Conservation and Correction Index (CACI)} --- quantifies whether healthy regions are preserved while anomalous regions are selectively corrected using structural similarity measures.
\end{itemize}
These indices complement Dice and AUROC by characterising pseudo-healthy reconstruction behaviour rather than lesion localisation alone. In a multi-reader study involving radiologists, stronger normative reconstruction metrics were associated with improved perceived plausibility and encouraging generalisation across unseen pathologies~\citep{Bercea2025}. Together, these findings suggest that evaluation of UAD models should consider both detection accuracy and the quality of normative modelling. Additionally, the absence of lesion-wise evaluation metrics limits assessment of the clinical utility of UAD models for sparse pathologies. Future evaluations should incorporate lesion-wise detection metrics (such as lesion-wise F1 score or false positive lesion count) to determine whether models with poor boundary segmentation can nonetheless reliably flag anomalous lesions.

Benchmarks that incorporate rare or heterogeneous pathologies (e.g., NOVA) may further stress-test generalisation beyond widely used datasets such as BraTS or MSSEG.

\paragraph{Clinical implications.}
To bridge the gap between retrospective benchmark performance and real-world clinical utility, future generative UAD frameworks should complement anomaly maps and pseudo-healthy reconstructions with estimates of predictive uncertainty. By incorporating uncertainty estimation, for example through Bayesian neural networks~\citep{Kohl2018} or diffusion-based sampling uncertainty~\citep{Wolleb2022}, algorithms may provide voxel-wise confidence scores alongside pseudo-healthy reconstructions. Factoring model uncertainty into the anomaly scoring process may help mitigate false positives, thereby improving the safety and reliability of generative UAD as a clinical triage tool.

Moreover, clinical translation will require greater emphasis on interpretability. While reconstruction-based generative UAD provides intuitive visual outputs through residual maps, further work is needed to determine whether these outputs reliably correspond to clinically meaningful abnormalities and can support clinical interpretation. Moving forward, the field should embrace multimodal AI architectures that integrate non-imaging clinical metadata (e.g., patient demographics, genomics, or clinical history) alongside multi-contrast MRI~\citep{Jiang2025}. This integration is crucial to construct a more robust, patient-specific definition of normative anatomy, particularly to account for age-related structural changes. To seamlessly integrate UAD outputs into the clinical workflow, imaging foundation models and large language models could be leveraged to automatically translate generative anomaly localisations into structured, contextualised preliminary reports~\citep{Zhang2026}.

Unsupervised generative models may be particularly suited to broad anomaly detection or triage settings, especially where pathology-specific annotations are unavailable. Their pseudo-healthy reconstructions provide interpretable contrastive visualisations that may complement supervised segmentation models. However, it is critical to clearly separate retrospective benchmark performance from actual clinical utility. The currently available evidence is insufficient to support independent or routine clinical use. To bridge this translational gap, prospective, multi-site reader studies are needed to determine whether pseudo-healthy reconstructions improve detection confidence, workflow efficiency, or longitudinal consistency under real-world clinical conditions. Ultimately, demonstrating clinical safety and efficacy remains a strict prerequisite for securing necessary regulatory approvals, such as CE marking or FDA clearance, before any of these tools can be deployed in routine medical practice.

\paragraph{Limitations.}
This review has several limitations. First, although six information sources were searched (PubMed, Web of Science, ScienceDirect, Springer Nature Link, IEEE Xplore, and ArXiv), some relevant studies may have been missed. Restricting ArXiv searches to the \textit{Computer Science (cs)} category may have excluded biomedical preprints. Second, the cut-off date (17 December 2025) introduces temporal bias in a rapidly evolving field. Third, quantitative synthesis relied primarily on Dice, AUROC, and AUPRC. Dice disproportionately penalises small lesions such as MS or WMH, AUROC can obscure class imbalance, and AUPRC introduces its own biases and is not inherently superior~\citep{McDermott2025}. Other measures were reported inconsistently and could not be systematically compared. Fourth, evaluation strategies varied substantially across studies, particularly regarding residual threshold selection. Finally, conventional metrics do not fully capture the normative modelling objective of reconstruction-based UAD~\citep{Bercea2025}, underscoring the need for task-specific evaluation frameworks.

\section{Conclusion}
In this systematic scoping review, we compared generative AI-based methods for anomaly detection and segmentation in brain MRI, focusing on their ability to model healthy anatomy and detect deviations. None of the included studies solved the challenge across all pathologies. In detection, some methods achieved high AUROC values ($>0.9$), but performance was typically pathology- or dataset-specific. No generalisable detection framework has yet emerged across pathologies and datasets. Segmentation remains particularly challenging: Dice scores for large lesions were moderate (often between 0.6–0.8), whereas performance for small or sparse lesions frequently fell below 0.3 and occasionally below 0.1.
We categorised studies by pathology and summarised their main contributions (Table~\ref{tab:method_summary}). Following PRISMA guidelines, we provided a transparent and reproducible synthesis, identifying consistent performance patterns across pathologies and highlighting emerging innovation.
In summary, unsupervised generative models provide a valuable, annotation-free strategy for detecting and visualising neuroimaging anomalies. However, performance remains limited for small or sparse lesions, and these methods do not yet match supervised baselines. Future work should prioritise anatomy-aware architectures, standardised multi-pathology benchmarks, and prospective reader studies to establish whether pseudo-healthy reconstructions can translate into clinically meaningful support, such as in research and follow-up exams.

\section*{Data and materials availability}  
All extracted data, screening records, and analysis code (including scripts for figure generation 
and an interactive table summarising tables \ref{tab:method_summary}, \ref{tab:detection_metrics} and \ref{tab:dice_summary}) are openly available in the supporting GitHub repository\footnote{https://github.com/youwanM/Unsupervised-Deep-Generative-Models-for-Anomaly-Detection-in-Neuroimaging}. The full PRISMA-ScR checklist is provided in the Supplementary Materials.  

\section*{Competing interests and funding} 
The authors declare no conflicts of interest. No review protocol was prepared, and this review was not preregistered in PROSPERO or any other registry. This work was conducted as part of a CIFRE doctoral programme. No specific additional funding was received for this scoping review.
Large language models (LLMs) were used exclusively for writing assistance, editing, and formatting. They did not contribute to study design, methodology, data analysis, or interpretation, and therefore did not affect the originality or scientific rigour of this work. 
\paragraph{CRediT authorship contribution statement} 
\textbf{YM}: Conceptualisation, Methodology, Investigation, Data curation, Writing – original draft. 
\textbf{EB}: Methodology, Supervision, Writing – review \& editing. 
\textbf{SL}: Methodology, Supervision, Writing – review \& editing. 
\textbf{EF}: Supervision, Writing – review \& editing. 
\textbf{FG}: Methodology, Validation, Supervision, Writing – review \& editing.
\noindent All authors read and approved the final manuscript.

\includepdf[pages=-,pagecommand=\section*{Supplementary Material}, width=0.8\linewidth]{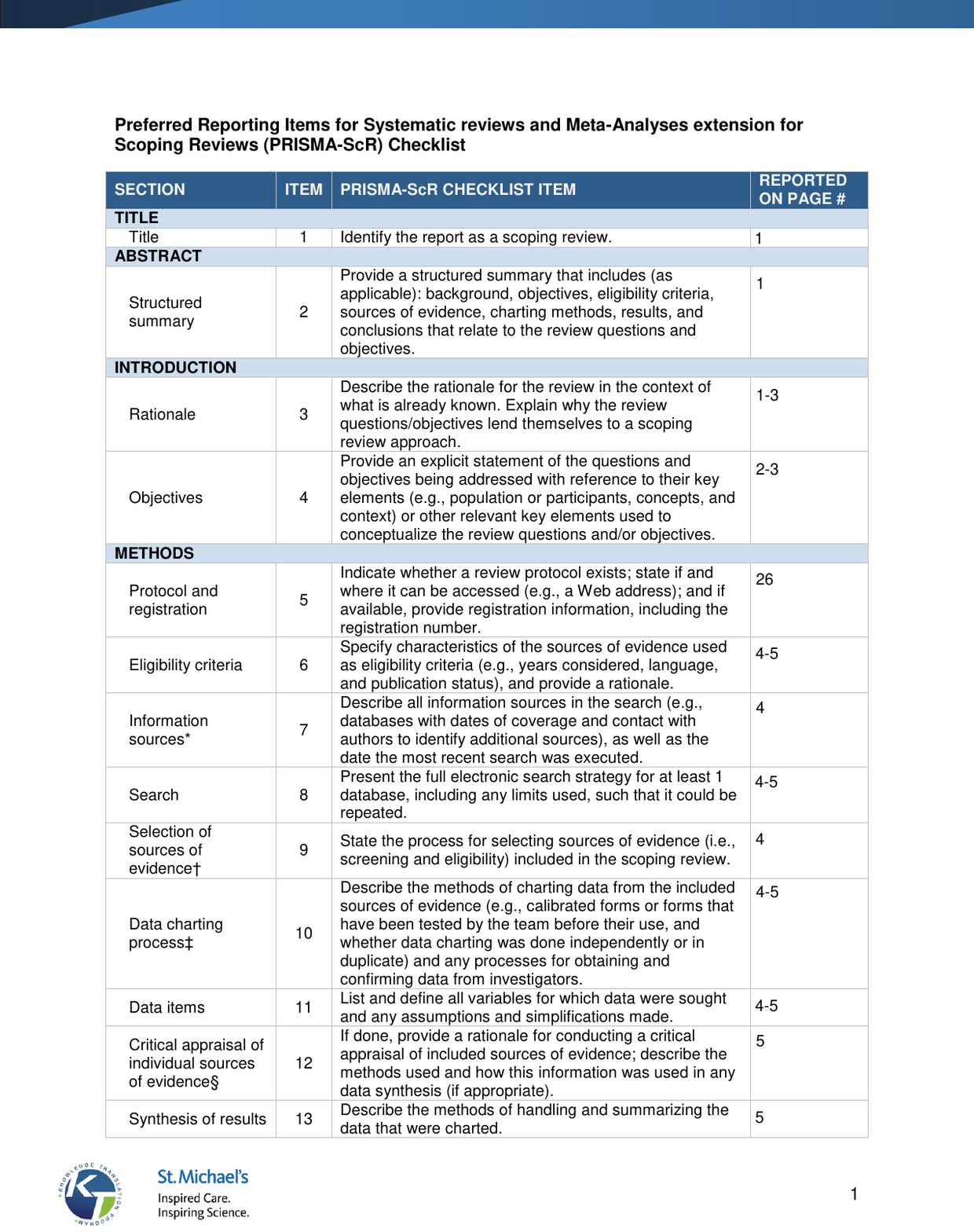}

\bibliographystyle{elsarticle-num-names}
\bibliography{references,rayyan}

\end{document}